\documentclass[journal]{IEEEtran}
\usepackage{graphicx}
\usepackage{amsmath}
\usepackage{amssymb}
\usepackage{subfigure}
\usepackage{stfloats}
\usepackage{algorithm}
\usepackage{algorithmicx}
\usepackage{algpseudocode}
\usepackage{url}
\usepackage{tabularx}
\usepackage{array}
\usepackage{multirow}
\usepackage{color}
\usepackage{booktabs}
\usepackage{enumerate}
\usepackage{breqn}
\usepackage{indentfirst}
\usepackage{bm}

\begin{document}
\thispagestyle{empty}

\onecolumn
{\centering
	This paper has been accepted for publication in IEEE Transactions on Instrumentation and Measurement.
	\vspace{5mm}
	
	DOI: 10.1109/TIM.2019.2949318\\
	IEEE Explore: \url{https://ieeexplore.ieee.org/document/8882497}
	
	\vspace{1cm}
	
	Please cite the paper as:\\
	\vspace{5mm}
	J. Wen, X. Zhang, H. Gao, J. Yuan, Y. Fang, ``CAE-RLSM: Consistent and efficient redundant line segment merging for online feature map building'', \emph{IEEE Transactions on Instrumentation and Measurement}, vol. 69, no. 7, pp. 4222-4237, 2020.
}

\twocolumn
\newpage
\setcounter{page}{1}

\title{CAE-RLSM: Consistent and Efficient Redundant Line Segment Merging for Online Feature Map Building}

\author{\IEEEauthorblockN{Jian Wen$^{1}$, Xuebo Zhang$^{1,2\dagger}$, Haiming Gao$^{1}$, Jing Yuan$^{1}$, and Yongchun Fang$^{1}$}
\thanks{This work is supported in part by National Key Research and Development Project under Grant 2018YFB1307503, in part by National Natural Science Foundation of China Under Grant U1613210 and 91848203, in part by Tianjin Science Fund for Distinguished Young Scholars, and  in part by Tianjin Natural Science Foundation under Grant 19JCYBJC18500, in part by the Open Project of the Key Laboratory of Industrial IoT and Networked Control Ministry of Education under Grant 2018FF07.}
\thanks{$^{1}$Jian Wen, Xuebo Zhang, Haiming Gao, Jing Yuan and Yongchun Fang are with the Institute of Robotics and Automatic Information System (IRAIS), Tianjin Key Laboratory of Intelligent Robotics (TJKLIR), Nankai University, Tianjin, P. R. China, 300350 (e-mail: zhangxuebo@nankai.edu.cn; wenjian@mail.nankai.edu.cn; ghm@mail.nankai.edu.cn; nkyuanjing@gmail.com; fangyc@nankai.edu.cn).}
\thanks{$^{2}$Xuebo Zhang is also with the Key Laboratory of Industrial IoT and Networked Control, Ministry of Education, Chongqing
University of Posts and Telecommunications, Chongqing 400065, China.}
\thanks{$\dagger$: Corresponding author.}
}

\maketitle

\begin{abstract}
In order to obtain a compact line segment-based map representation for localization and planning of mobile robots, it is necessary to merge redundant line segments which physically represent the same part of the environment in different scans. In this work, a consistent and efficient redundant line segment merging approach (CAE-RLSM) is proposed for online feature map building. The proposed CAE-RLSM is composed of two newly proposed modules: one-to-many incremental line segment merging (OTM-ILSM) and multi-processing global map adjustment (MP-GMA). Different from state-of-the-art offline merging approaches, the proposed CAE-RLSM can achieve real-time mapping performance, which not only reduces the redundancy of incremental merging with high efficiency, but also solves the problem of global map adjustment after loop closing to guarantee global consistency. Furthermore, a new correlation-based metric is proposed for the quality evaluation of line segment maps. This evaluation metric does not require manual measurement of the environmental metric information, instead it makes full use of globally consistent laser scans obtained by simultaneous localization and mapping (SLAM) systems to compare the performance of different line segment-based mapping approaches in an objective and fair manner. Comparative experimental results with respect to a mean shift-based offline redundant line segment merging approach (MS-RLSM) and an offline version of one-to-one incremental line segment merging approach (O$^{2}$TO-ILSM) on both public datasets and self-recorded dataset are presented to show the superior performance of CAE-RLSM in terms of efficiency and map quality in different scenarios.
\end{abstract}


\IEEEpeerreviewmaketitle

\section{Introduction}
\IEEEPARstart{M}{AP} building is a key module for autonomous mobile robots since both localization and planning usually depend on the map information of the environment \cite{1,2}. One typical map representation is the grid map \cite{3,4}, wherein the value of each cell represents the probability of being occupied by obstacles. However, the required memory of the grid map grows rapidly with the increase of the environmental size. Therefore, the representation of the environment by features such as line segments becomes a popular alternative to the grid map \cite{6,10,11,26}. Compared with the grid map, the line segment map is more compact, consuming less memory and scaling better with the environmental size. Furthermore, the line segment map does not split the environment into grids and thus it does not suffer from discretization problems \cite{27}.

An important preparation before building a line segment map is extracting line segments from discrete and ordered range sensor data. The line segment extraction algorithms have been studied well, and there are many classical algorithms such as \emph{Iterative End Point Fit} algorithm, \emph{Hough Transform} algorithm and so on \cite{9, 27}. A simple way to build a local or global line segment map is to transform the original line segments from the local coordinates to the global coordinates and pile them up. The local map can be regarded as a subset containing the line segments extracted from several consecutive laser scans, and the global map is the total set containing all original line segments. However, because of the errors in localization or perception, those line segments physically representing the same part of the environment are not perfectly overlapped in both local and global maps. This situation is called redundancy and inconsistency of map information, which makes it difficult to fulfill subsequent tasks such as localization and planning. Therefore, it is necessary to merge redundant line segments in the original line segment map to obtain a non-redundant and globally consistent map. Existing redundant line segment merging approaches can be classified into two categories: offline approaches \cite{6,13,14} and online approaches \cite{10,19}. Offline approaches collect all laser scans and register them in advance so that they are globally consistent. Then a variety of clustering algorithms are adopted to classify these line segments extracted from the registered laser scans and merge them. The redundancy of line segment maps obtained by offline approaches is usually low, but the post-processing mechanism makes them essentially impossible to apply to the real-time systems, such as a feature-based graph SLAM system. In contrast, online approaches receive laser scans in real time and correct robot poses by scan matching. Then line segments extracted from the currently received laser scan are updated to the global map by incremental merging. This incremental processing mechanism can ensure the real-time performance, but it is difficult to handle the problem of global map adjustment after loop closing \cite{12}. In summary, the primary challenge in design of a redundant line segment merging approach is to simultaneously ensure high efficiency and global consistency.

Furthermore, how to evaluate the quality of line segment maps is a key issue. The fairest way is to compare the obtained maps with the actual environmental metric information. However, manual measurement of the environmental metric information usually requires a lot of labor and it is almost impractical in some cases. Therefore, most of past researches only judge if the obtained line segment maps are redundant or not by visual inspection \cite{10,11,13,14,25,26}. Some researchers suggest using the \emph{reduction rate}, i.e., the ratio of the reduced number of line segments after merging to the number of the original line segments before merging, to evaluate the performance of redundant line segment merging approaches \cite{5,6}. However, it is not that the higher the reduction rate, the better the map. A high reduction rate may also mean serious loss of map information, because some physically different line segments may have been erroneously merged together. In addition, the reduction rate cannot directly reflect the redundancy of line segment maps, because even the high reduction rate cannot ensure that there are no redundant line segments in the map. Therefore, how to objectively and fairly evaluate the quality of line segment maps is yet to be studied further.

\begin{figure}[t]
    \centering
    \includegraphics[scale=0.28]{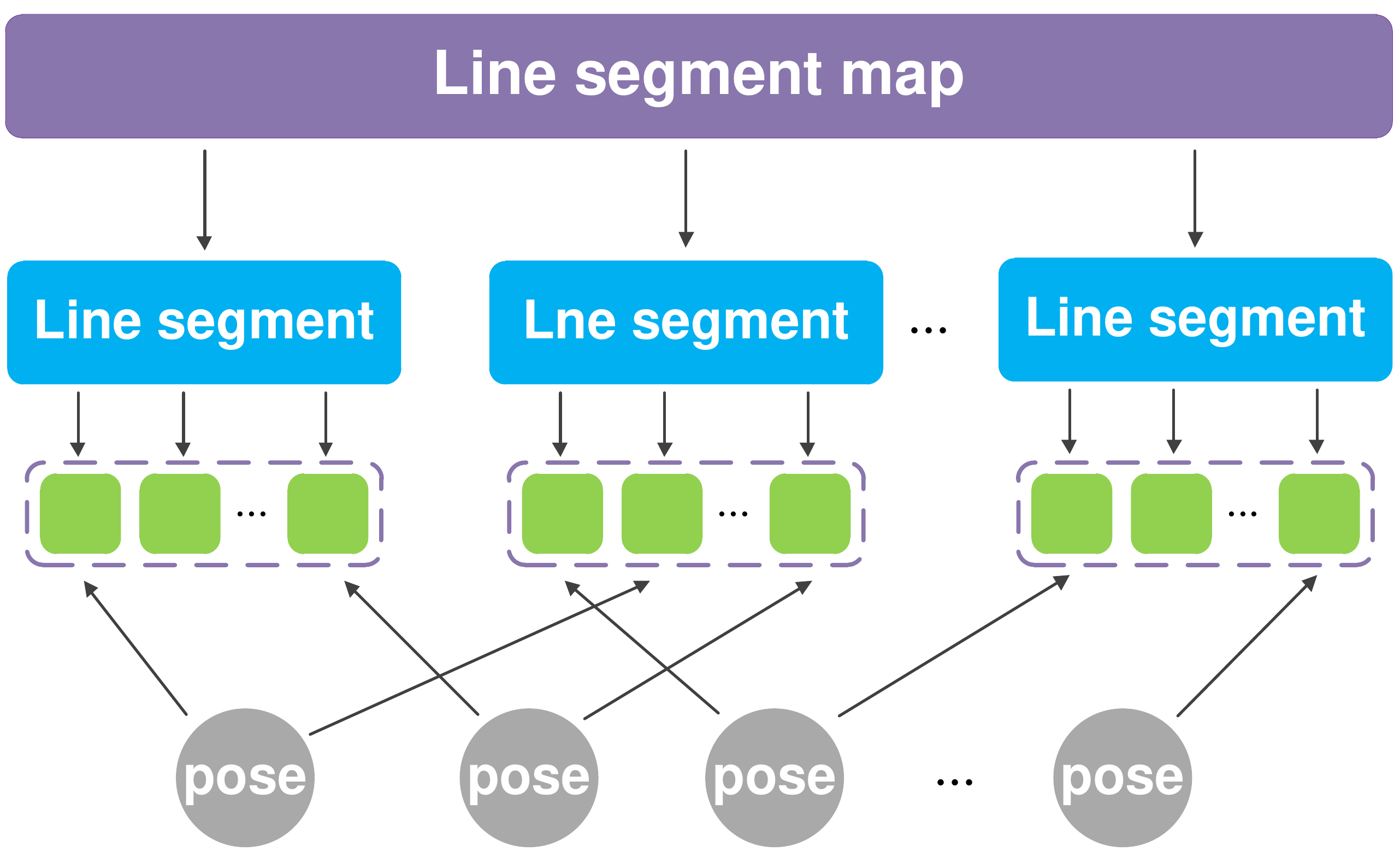}
    \caption{Relationship between the final line segment map (purple), line segments in the final map (blue), subsets of original line segments (green) and robot poses (dark gray).}
    \label{frame}
\end{figure}

Motivated by these challenges, a consistent and efficient redundant line segment merging approach called CAE-RLSM is newly proposed. The proposed CAE-RLSM is composed of two newly proposed modules: one-to-many incremental line segment merging (OTM-ILSM) and multi-processing global map adjustment (MP-GMA). The final line segment map is regarded as a global set and each line segment in the final map is regarded as a subset. Each subset contains all original line segments that make up the final line segment corresponding to this subset. The correspondence between the original line segments and the corresponding final line segment is recorded during incremental merging, and all original line segments are retained after merging. These original line segments are extracted from raw laser scans by a seeded region growing-based line segment extraction algorithm \cite{18} and associated with the corresponding robot poses. After loop closing, all original line segments are updated to the world coordinates according to the optimized poses. Finally, all original line segments in each subset are re-merged into a new line segment to update the final map. This step reuses the previous recorded correspondence between the original and final line segments instead of executing incremental merging from scratch, which significantly improves efficiency while guaranteeing global consistency. The relationship between the above components can be seen in Fig. \ref{frame}. It should be emphasized that the subsets of original line segments corresponding to the final line segments in the final map are disjoint. In addition, it should be noted that CAE-RLSM is independent of loop closure detection and pose optimization. Therefore, it can be embedded as a stand-alone module in standard graph-based SLAM systems.

Furthermore, a new correlation-based metric is proposed to evaluate the quality of line segment maps. Globally consistent laser scans obtained by SLAM systems are regarded as ground truth, which are collected and rasterized using a Gaussian kernel to construct a lookup table. Each line segment in the final map is discretized into pixels using the Bresenham algorithm \cite{33}. And then these pixels are projected to match the corresponding value in the lookup table. The quality of the final map is computed by adding up the scores of all line segment pixels. To the best of our knowledge, we are the first to propose such correlation-based metric for quality evaluation of line segment maps.

The main contributions of this paper can be summarized as follows:

(1) A novel redundant line segment merging approach called CAE-RLSM is proposed for online feature map building, which consists of two new modules OTM-ILSM and MP-GMA. Different from existing online and offline approaches, CAE-RLSM not only reduces the redundancy of incremental merging, but also solves the problem of global map adjustment after loop closing. In other words, CAE-RLSM ensures both high efficiency and global consistency to obtain a compact line segment-based map.

(2) A new evaluation metric based on correlation is proposed for quality evaluation of line segment maps. This evaluation metric does not require manual measurement of the environmental metric information, but makes full use of globally consistent laser scans obtained by SLAM systems to objectively and fairly compare the performance of different line segment-based mapping approaches.

We begin this paper with a review of related work in Section II. The proposed CAE-RLSM is detailed in Section III. Section IV describes the evaluation metrics and Section V presents the comparative experimental results and analysis. Finally, this paper is concluded in Section VI.

\section{Related Work}
Map building has received much research attention in recent years due to its value in abundant applications for mobile robots, such as localization \cite{2,7,17}, pose tracking \cite{22}, place recognition \cite{16} and navigation \cite{8}. In \cite{7}, Liu et al. propose a feasible solution for room-level localization through a structured kernel sparse coding model. The work \cite{17} presents a hybrid global localization approach for indoor mobile robots with ultrasonic sensors. In \cite{16}, Zhuang et al. design a 3D laser-based place recognition system for a mobile robot to recognize complex indoor scenes.

Because of the compact representation and low memory, line segment feature has been widely used in mobile robots localization, mapping and navigation \cite{23,24,25,26,28,30,31}. In order to obtain a compact and efficient map representation based on line segments, it is necessary to merge redundant line segments which physically represent the same part of the environment such as a long wall. In \cite{13}, Amigoni and Vailati propose a greedy offline approach for redundant line segment merging. In this work \cite{13}, they first consider the longest line segment in the whole map and then define a strip centred on this longest line segment with a width of $2\epsilon$. Other line segments whose endpoints fall within the strip are merged together with this longest line segment. In \cite{6}, Sarkar et al. propose an offline redundant line segment merging approach, which is mainly based on the mean shift clustering algorithm and thus called MS-RLSM in this paper. In the work \cite{6}, a three-layer architecture for redundant line segment merging is proposed. First, all original line segments are clustered into angular clusters according to their direction by the mean shift clustering algorithm; then, line segments belonging to the same angular cluster are further clustered into spatial clusters based on their spatial proximity; finally, the line segments within the same spatial cluster are merged together. Although above offline approaches can significantly reduce the redundancy of final maps, they have poor running time performance, which makes it difficult to apply to real-time mapping systems.

In \cite{19}, Kuo et al. propose the naive one-to-one incremental line segment merging approach (OTO-ILSM), which is applied in a Rao-Blackwellized Particle Filters (RBPF) based SLAM system. In this work \cite{19}, for each line segment extracted from the currently received laser scan, a complete search through the global map is performed. Once the closest match is found, they regard the two line segments belonging to the same part of the environment and merge them together. However, OTO-ILSM has limitations in both redundancy reduction and global consistency. Firstly, when there is more than one line segment in the global map that matches with a currently extracted line segment, only one of them will be merged, resulting in other redundant line segments being left in the global map. In particular, when a robot wraps around a large circle and returns to the start position, using OTO-ILSM cannot merge the line segments extracted at the start and end positions in a single loop together with the currently extracted line segment, since the currently extracted line segment will be only merged with one of them. Secondly, it is essentially impossible for this incremental processing mechanism to adjust global line segment map according to the optimized poses after loop closing, since robot poses are only associated with original line segments extracted from raw laser scans while line segments in the global map are not associated with robot poses. A simple way to address this problem is to execute incremental merging from scratch after loop closing. However, this in turn reduces the efficiency of incremental merging.

In order to obtain non-redundant and globally consistent line segment maps in real-time, in this work a novel redundant line segment merging approach called CAE-RLSM is proposed. To reduce the redundancy of incremental merging, a one-to-many incremental line segment merging module (OTM-ILSM) is designed. The correspondence between the original line segments extracted from raw laser scans and the final line segments in the global map is recorded during incremental merging, and all original line segments are retained after merging. This correspondence is reused for global map adjustment after loop closing instead of executing incremental merging from scratch, which significantly improves efficiency while guaranteeing global consistency. The proposed CAE-RLSM is compared with MS-RLSM \cite{6} and OTO-ILSM \cite{19} on both public datasets and self-recorded dataset to demonstrate that CAE-RLSM can achieve superior performance in terms of efficiency and map quality in different scenarios.

Furthermore, it is necessary to evaluate the quality of line segment maps to compare the performance of different line segment-based mapping approaches. In order to quantitatively evaluate the quality of line segment maps without ground truth, several evaluation metrics have been designed \cite{5,6,21}. In \cite{6}, Sarkar et al. use the number and length of line segments in the final map to evaluate the quality of obtained maps. They believe that the fewer the number of line segments in the final map, the more compact the line segment map. Similarly, Amigoni and Quattrini Li propose the reduction rate, i.e., the ratio of the reduced number of line segments after merging to the number of the original line segments before merging, to compare the efficiency of different redundant line segment merging approaches \cite{5}. But in fact, fewer line segments in the final map or higher reduction rate may also mean serious loss of map information, since some physically different line segments are merged together by mistake, and the final map cannot preserve the structural information of the environment well. Therefore, it is not objective and fair to evaluate the quality of line segment maps only by the number of line segments in the final map. However, the quality metric presented in \cite{5} is a comprehensive distance metric between the original line segments and the final one after merging. Specifically, since each final line segment in the final map has its corresponding original line segments before merging, the weighted summation of the distance deviations and angular deviations between the final line segments and the corresponding original line segments is defined as the loss of map information deriving from redundant line segment merging process.

In this work, we borrow the idea of correlative scan matching algorithm (CSM) \cite{20} and propose a new correlation-based metric for quality evaluation of line segment maps. Globally consistent laser scans obtained by SLAM systems are regarded as ground truth and rasterized using a Gaussian kernel to construct a lookup table. This lookup table can be regarded as the environmental model. Each line segment in the final map is discretized into pixels using the Bresenham algorithm. Finally, these pixels are projected and matched with the corresponding value in the lookup table to evaluate the quality of obtained maps. In addition to the newly proposed evaluation metric, we also introduce a simplified error metric to comprehensively evaluate the quality of line segment maps, which is based on the idea of the quality metric presented in \cite{5}. This error metric is defined as the average deviation of the barycenters of the original line segments to the corresponding final line segments, which represents the average deviation of the original line segments to the final map.

\section{CAE-RLSM}
\begin{figure}[t]
    \centering
    \includegraphics[scale=0.24]{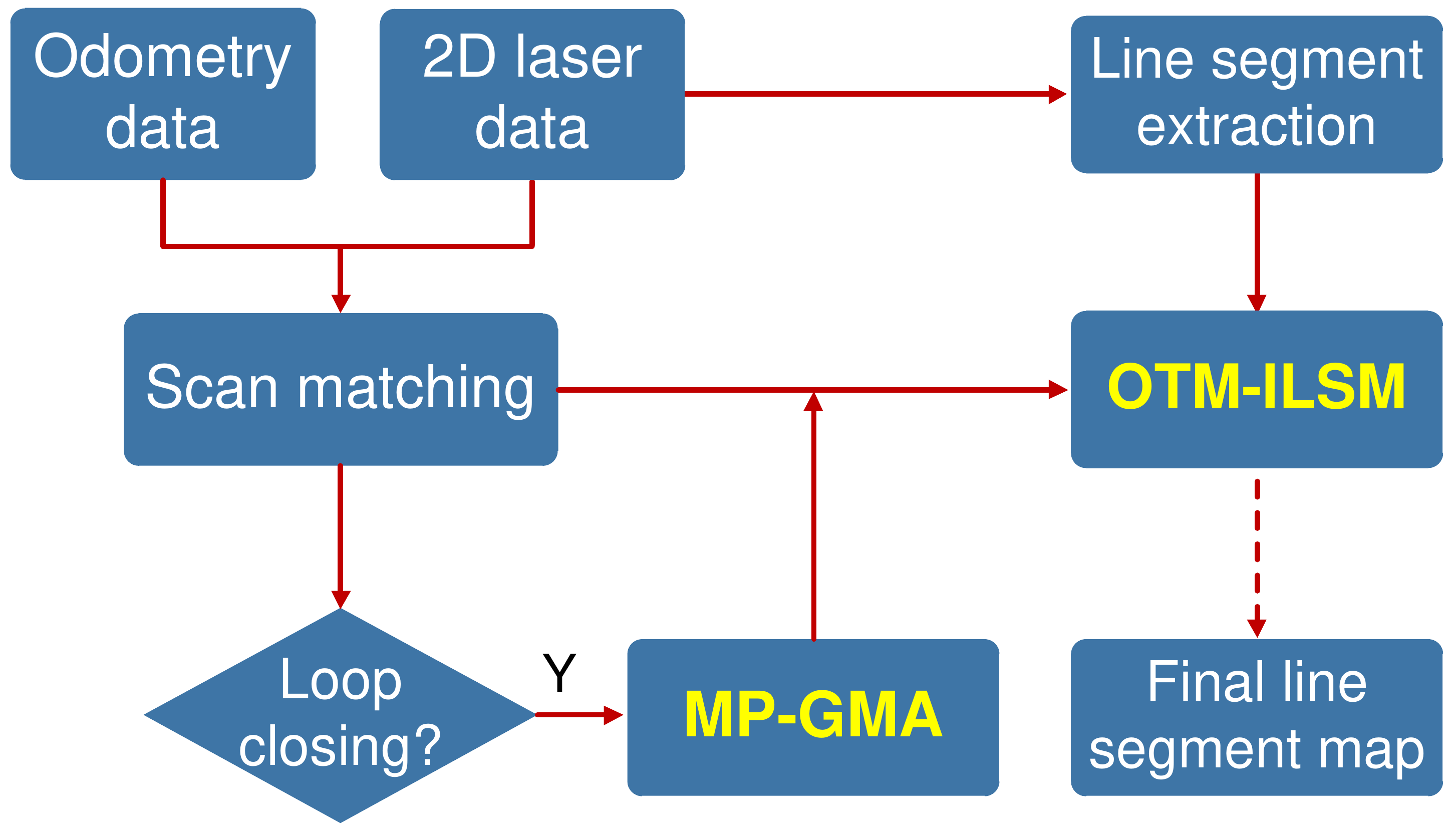}
    \caption{System framework. The proposed CAE-RLSM is composed of one-to-many incremental line segment merging (OTM-ILSM) and multi-processing global map adjustment (MP-GMA). When new sensor data are received, the modules of scan matching and loop detection are called successively. If loop closing is executed, the modules of MP-GMA and OTM-ILSM are called successively. Otherwise, only OTM-ILSM is called.}
    \label{flowchart}
\end{figure}

This section introduces the framework and detailed procedures of CAE-RLSM. As shown in Fig. \ref{flowchart}, CAE-RLSM updates the global line segment map when new sensor data are received. Original line segments are extracted from the raw laser scan, and scan matching algorithms are used to correct the robot pose. If the received laser scan triggers the loop closure conditions, loop closing is called to update all robot poses, and all original line segments are updated and re-merged to update the global line segment map by multi-processing global map adjustment (MP-GMA) module. Otherwise, the one-to-many incremental line segment merging (OTM-ILSM) module is called to merge the currently extracted line segments with the last global line segment map. It should be noted that MP-GMA always works together with OTM-ILSM to eliminate the redundancy of the final line segment map, thus they are two closely integrated modules.

\subsection{Notations}
A single line segment $l$ is represented by its two endpoints $\left( {{x_i},{y_i}} \right),i = 1,2$. The heading of the line segment is defined as the direction vector of the start point to the end point and is normalized into the range of \(\left( { - \pi ,\pi } \right]\). ${l^G}$ and ${l^L}$ represent the line segment expressed in the global and local coordinates, respectively. ${S_t} = \left\{ {{l_1},{l_2},...,{l_n}} \right\}$ denotes the set of line segments extracted from $t$-th laser scan. $S_t^G$ and $S_t^L$ represent the set of line segments expressed in the global and local coordinates, respectively.

The set ${C_t}$ contains all original line segments extracted from raw laser scans collected during the time interval $\left[ {0,t} \right]$. Let ${M_t} = \left\{ {l_1^G,l_2^G, \ldots ,l_k^G} \right\}$ denote the global line segment map at time $t$, and ${C_t}$ is explicitly denoted as ${C_t} = \left\{ {{c_1},{c_2}, \ldots ,{c_k}} \right\}$, where each subset ${c_i},i = 1,2, \ldots k$ contains all original line segments $\left\{ {l_{i1}^L,l_{i2}^L, \ldots } \right\}$ before merging that make up the corresponding line segment $l_i^G$ in ${M_t}$ after merging.

\subsection{Fusion Conditions and Fusion Algorithm}
According to \cite{12}, a general framework for designing a redundant line segment merging approach is usually structured with several \emph{fusion conditions} and a \emph{fusion algorithm}. The fusion conditions are used to detect several line segments which physically represent the same part of the environment, and these line segments are merged into a new line segment through the fusion algorithm. Furthermore, each line segment in the final map is usually assigned with a weight according to its length or the number of times it is updated. In this paper the latter is used.

\begin{figure}[t]
    \centering
    \includegraphics[scale=0.6]{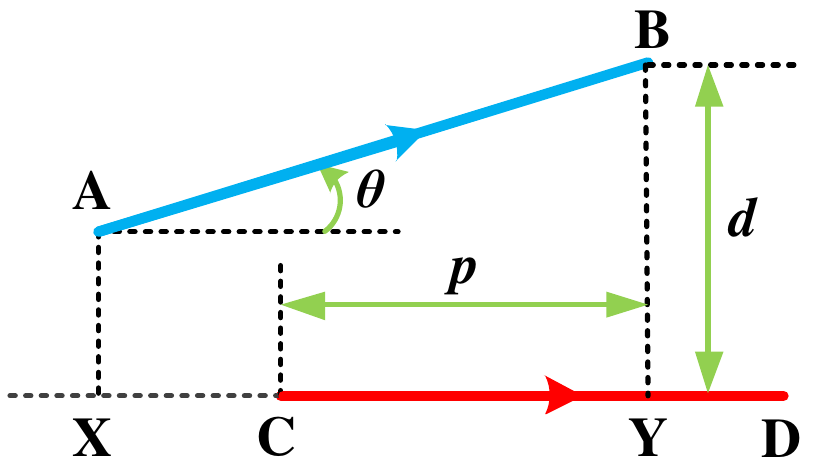}
    \caption{The heading deviation, separation distance and overlap between the line segment extracted from the currently received laser scan (AB) and the line segment from the global map (CD) are represented by $\theta$, $d$ and $p$, respectively. The projections of the endpoints of AB to CD are denoted by X and Y.}
    \label{fusioncondition}
\end{figure}

In \cite{6}, the concept of spatial proximity based on lateral separation and longitudinal overlap is proposed. This spatial proximity is suitable for offline approaches because the weight of all original line segments extracted from raw laser scans is equal. But for online approaches, the weight of the line segments from the latest global line segment map ${M_{t - 1}}$ should be larger than the weight of the currently extracted line segments, since those line segments from ${M_{t - 1}}$ may have been updated many times. Therefore, it is not appropriate to treat such two kinds of line segments as equal for online approaches. In this paper, the following three measurements are considered to determine whether two line segments, ${l_M}$ and $l_S^G$, one from ${M_{t - 1}}$ and the other from $S_t^G$, are spatially close or not:

\begin{enumerate}[(a)]
\item Heading deviation, $\theta$: the difference between the headings of $l_S^G$ and ${l_M}$.
\item Separation distance, $d$: the maximum distance of the endpoints of $l_S^G$ to ${l_M}$.
\item Overlap, $p$: the length of the overlap between ${l_M}$ and the projection of $l_S^G$ to ${l_M}$.
\end{enumerate}
If and only if the following three conditions are all satisfied, ${l_M}$ and ${l_S^G}$ are considered to be spatially close to each other:
\begin{equation}
    \left| {{\textsl{normAngle}}\left( {{\theta ^M} - {\theta ^{SG}}} \right)} \right| \le {\theta _{{\textup{max}}}},
\end{equation}
\begin{equation}
    {\textup{max}}\left\{ {\frac{{\left| {{A^M}x_i^{SG} + {B^M}y_i^{SG} + {C^M}} \right|}}{{\sqrt {{{({A^M})}^2} + {{({B^M})}^2}} }}} \right\} \le {d_{{\textup{max}}}},i = 1,2,
\end{equation}
\begin{equation}
    \sqrt {{{\left( {x_1^{EG} - x_2^{EG}} \right)}^2} + {{\left( {y_1^{EG} - y_2^{EG}} \right)}^2}}  \ge {p_{\textup{min}}},
\end{equation}
where ${\theta ^M}$ and ${\theta ^{SG}}$ denote the headings of ${l_M}$ and $l_S^G$, respectively. $({A^M},{B^M},{C^M})$ denotes the three parameters of the general form of the equation of ${l_M}$. $(x_1^{SG},y_1^{SG})$ and $(x_2^{SG},y_2^{SG})$ denote the endpoints of $l_S^G$. $(x_1^{EG},y_1^{EG})$ and $(x_2^{EG},y_2^{EG})$ are the endpoints of the overlap between ${l_M}$ and the projection of $l_S^G$ to ${l_M}$, and ${\theta _{\textup{max}}}$, ${d_{\textup{max}}}$ and ${p_{\textup{min}}}$ are three threshold parameters. Since the angles are not Euclidean, the function $\textsl{normAngle}\left(  \bullet  \right)$ is used to re-normalize them after every subtraction. The diagram of above three measurements is shown in Fig. \ref{fusioncondition}.

Since the direction of the line segment is considered, if a robot first moves from left to right along a wall and then moves from right to left along the same side of the wall, the two line segments acquired along the two moving directions will be correctly merged together. It should be emphasized that the measurement of heading deviation $\theta$ is not a redundant parameter even though the other two measurements of separation distance $d$ and overlap $p$ are strongly correlated with it. Considering the following situation, a robot moves around a thin wall and acquires two parallel and very close line segments in two sides of the wall. Apparently, these two line segments represent different environmental structures. However, if we only consider the separation distance $d$ and overlap $p$ and use a bit slack parameter settings, these two line segments may be judged as a pair of redundant line segments and erroneously merged together. If the direction of the line segment is considered, these two line segments will be judged as two irrelevant line segments since they have opposite directions. Namely, it is not enough to use only the measurements of $d$ and $p$ to find associated line segments. Furthermore, the calls to the above three measurements are sequential. If the heading deviation $\theta$ between two line segments exceeds the angular threshold, the two line segments are considered irrelevant, and the separation distance $d$ and overlap $p$ between them will not be calculated. In other words, the heading deviation $\theta$ is used for coarse screening while the separation distance $d$ and overlap $p$ are used for fine screening. Such hierarchical screening aims to increase the computational efficiency since the calculation of the last two measurements needs more computing resources.

For each line segment in $S_t^G$, the associated line segments in ${M_{t - 1}}$ satisfying the fusion conditions are selected to constitute a set $\left\{ {l_i^G} \right\},i = 1,2, \ldots ,N$, which contains line segments physically representing the same part of the environment. And the following equations are used to merge these line segments into a new global line segment ${\ell}$:
\begin{equation}
    {\bar x^{CG}} = \frac{{\sum\limits_{i = 1}^N {{w_i}x_i^{CG}} }}{{\sum\limits_{i = 1}^N {{w_i}} }},{\bar y^{CG}} = \frac{{\sum\limits_{i = 1}^N {{w_i}y_i^{CG}} }}{{\sum\limits_{i = 1}^N {{w_i}} }},\bar \theta  = \frac{{\sum\limits_{i = 1}^N {{w_i}{\theta _i}} }}{{\sum\limits_{i = 1}^N {{w_i}} }},
\end{equation}
\begin{dmath}
    \left\{ {\left( {{{\bar x}_1},{{\bar y}_1}} \right),\left( {{{\bar x}_2},{{\bar y}_2}} \right)} \right\} = \mathop {{\textup{argmax}}}\limits_{i \in N,j \in N} \left\{ {\sqrt {{{\left( {x_i^{E1} - x_j^{E2}} \right)}^2} + {{\left( {y_i^{E1} - y_j^{E2}} \right)}^2}} } \right\},
\end{dmath}
where $(x_i^{CG},y_i^{CG})$ and $({\bar x^{CG}},{\bar y^{CG}})$ denote the center points of $l_i^G$ and ${\ell}$, respectively. $\bar \theta $ denotes the heading of ${\ell}$ and $w$ is the weight of the line segment, which corresponds to the number of times the line segment is updated. The projections of the endpoints of $l_i^G$ to ${\ell}$ are denoted by $(x_i^{E1},y_i^{E1})$ and $(x_i^{E2},y_i^{E2})$, and the two farthest projections form the new endpoints of ${\ell}$, which are represented by $({\bar x_1},{\bar y_1})$ and $({\bar x_2},{\bar y_2})$.

\textsl{Remark 1:} Normally, the speed of the robot is not very fast and the inter-frame error of the pose estimation provided by scan matching module is actually small. Therefore, the same line segment features will be observed and extracted many times from places having typical linear structures. The line segments corresponding to these linear structures will be updated many times. Thus, it is reasonable to consider that these line segments are more stable and reliable and give them greater weight.

\subsection{Data Structure}
CAE-RLSM is based on a tree-like data structure, as shown in Fig. \ref{frame}. The root node contains the final line segment map, and each of its child nodes contains a final line segment marked with a unique index. Furthermore, the children of each child node contain the original line segments that make up the corresponding line segment in the final map after merging. These original line segments are extracted from raw laser scans and marked with the indices of corresponding robot poses. And these original line segments are also regarded as leaf nodes of the tree-like data structure. During the incremental merging, the child nodes of the root node may be pruned, but these leaf nodes will be retained to record the information of all original line segments. Based on this tree-like data structure, the correspondence between the original line segments and the corresponding line segment in the final map is recorded, and each original line segment is also associated with its corresponding robot pose, facilitating global line segment map adjustment after loop closing.

\subsection{One-to-many Incremental Line Segment Merging}
Incremental line segment merging updates the global line segment map through merging ${S_t^G}$ with ${M_{t - 1}}$. As mentioned before, OTO-ILSM has limitations in both redundancy reduction and global consistency. Therefore, we propose the OTM-ILSM (Alg. 1).

In Alg. 1, for each currently extracted line segment in ${S_t^G}$, a complete search through ${M_{t - 1}}$ is performed (Alg. 1: lines 2-5). Line segments in ${M_{t - 1}}$ which satisfy the fusion conditions are regarded as the associated line segments of the currently extracted line segment and moved from ${M_{t - 1}}$ to a empty set $A$ (Alg. 1: lines 6-11). And then these associated line segments are merged together with the currently extracted line segment to generate a new line segment $\ell$ using the fusion algorithm (Alg. 1: line 13). The new line segment $\ell$ is inserted into ${M_{t - 1}}$ with the minimum index of these associated line segments (Alg. 1: line 20). Finally, those currently extracted line segments in ${S_t^G}$ which are not associated with line segments in ${M_{t - 1}}$ are directly inserted into ${M_{t - 1}}$ to update ${M_{t - 1}}$ to ${M_t}$ (Alg. 1: lines 25-26).

\textsl{Remark 2:} For one currently extracted line segment $l$, the set $A$ in Alg. 1 contains all associated line segments in the global map which satisfy the fusion conditions with $l$, and the set $B$ records the corresponding indices of these associated line segments. Given a set of line segments, the function $\textsl{Index}\left(  \bullet  \right)$ returns the corresponding indices of these line segments. And the identifier \textsl{lastIndex} in Alg. 1 records the maximum index of line segments in the global line segment map ${M_{t - 1}}$.

\begin{algorithm}[t]
        \caption{One-to-many Incremental Line Segment Merging}
        \begin{algorithmic}[1]
            \Require ${C_{t - 1}}$, ${M_{t - 1}}$, $S_t^G$, $S_t^L$ and \textsl{lastIndex}
            \Ensure ${C_t}$, ${M_t}$ and \textsl{lastIndex}
            \State ${N_1} = {\textsl{Index}}\left( {S_t^G} \right) = {\textsl{Index}}\left( {S_t^L} \right),{N_2} = {\textsl{Index}}({M_{t - 1}})$
            \For{$i \in {N_1},l_i^G \in S_t^G,l_i^L \in S_t^L$}
                \State $A$ $\leftarrow \emptyset$
                \State $B$ $\leftarrow \emptyset$
                \For{$j \in {N_2},{l_j^G} \in {M_{t - 1}}$}
                    \If{$l_i^G$ and ${l_j^G}$ satisfy the fusion conditions}
                        \State add ${l_j^G}$ into \textsl{A}.
                        \State add the index of ${l_j^G}$ into the set $B$.
                        \State erase ${l_j^G}$ from ${M_{t - 1}}$.
                    \EndIf
                \EndFor
                \State $m$ $\leftarrow$ the minimum number in the set $B$.
                \State merge all line segments in the set $A$ with $l_i^G$ into $\ell$.
                \For{$k \in B$, ${c_{k}} \in {C_{t - 1}}$}
                    \If{$k \ne m$}
                        \State move all line segments in ${c_{k}}$ to ${c_m}$.
                        \State erase ${c_{k}}$ from ${C_{t - 1}}$.
                    \EndIf
                \EndFor
                \State insert $\ell$ into ${M_{t - 1}}$ with the index $m$.
                \State erase ${l_i^G}$ from ${S_t^G}$.
                \State erase ${l_i^L}$ from ${S_t^L}$.
            \EndFor
            \State ${N_3} = {\textsl{Index}}\left( {S_t^G} \right) = {\textsl{Index}}\left( {S_t^L} \right)$
            \For{$i \in {N_3},l_i^G \in S_t^G,l_i^L \in S_t^L$}
                    \State insert ${l_i^G}$ into ${M_{t - 1}}$ with the index \textsl{lastIndex}.
                    \State create a new subset in ${C_{t - 1}}$ with the index \textsl{lastIndex}.
                    \State add ${l_i^L}$ into the new subset.
                    \State \textsl{lastIndex} $\leftarrow$ \textsl{lastIndex} + 1.
                \EndFor
            \State ${M_t} = {M_{t - 1}}$
            \State ${C_t} = {C_{t - 1}}$
        \end{algorithmic}
\end{algorithm}

In order to record the correspondence between the original line segments and the corresponding line segment in the final map, the tree-like data structure detailed in the previous subsection is used. In this subsection, the root node is represented by ${C_{t - 1}}$, which can be regarded as a set containing all original line segments. And the child nodes of the root node are denoted by $\left\{ {{c_i}} \right\},i = 1,2, \ldots ,k$. Each child node ${c_i}$ is regarded as a subset containing the original line segments $\left\{ {l_{i1}^L,l_{i2}^L, \ldots } \right\}$ that make up the corresponding final line segment ${l_i^G}$ in ${M_{t - 1}}$ after merging, and these original line segments are regarded as leaf nodes of the tree-like data structure. It should be noted that the indices of the subsets $\left\{ {{c_i}} \right\}$ are consistent with the indices of the final line segments in ${M_{t - 1}}$. During the complete search through ${M_{t - 1}}$, whenever a currently extracted line segment is merged together with its associated line segments in ${M_{t - 1}}$, the child nodes corresponding to these associated line segments are removed from $C_{t-1}$, and their leaf nodes are linked to a newly created child node whose index is the minimum index of these associated line segments (Alg. 1: lines 12-19). Finally, for each currently extracted line segment in ${S_t^G}$ which is not associated with line segments in ${M_{t - 1}}$, we create a new child node in ${C_{t - 1}}$ and link the currently extracted line segment with the new child node (Alg. 1: lines 27-28).

Compared with OTO-ILSM, the advantages of OTM-ILSM can be summarized as follows. Firstly, when there is more than one line segment in the global map matched with a currently extracted line segment, the currently extracted line segment will be merged together with all of them, reducing the redundancy of incremental merging. In particular, when the robot wraps around a large circle and returns to the start position, using OTM-ILSM can merge the line segments extracted at the start and end positions in a single loop together with the currently extracted line segment. Secondly, all original line segments are retained after incremental merging, and the correspondence between the original line segments and the corresponding line segment in the final map is recorded. This correspondence is reused for global map adjustment after loop closing instead of executing incremental merging from scratch, which significantly improves efficiency.

\subsection{Multi-processing Global Map Adjustment}
\begin{algorithm}[t]
        \caption{Multi-processing Global Map Adjustment}
        \begin{algorithmic}[1]
            \Require ${C_{t - 1}}$ and ${M_{t - 1}}$
            \Ensure ${M_{t - 1}}$
            \State ${N_4} = {\textsl{Index}}\left( {{C_{t - 1}}} \right) = {\textsl{Index}}\left( {{M_{t - 1}}} \right)$
            \For{$i \in {N_4},{l_i^G} \in {M_{t - 1}},{c_i} \in {C_{t - 1}}$}
                \State $D$ $\leftarrow \emptyset$
                \State ${N_5} = {\textsl{Index}}\left( {{c_i}} \right)$
                \For{$j \in {N_5},{l_j^L} \in {c_i}$}
                    \State update $l_j^L$ according to the optimized pose to $l_j^G$.
                    \State add $l_j^G$ into the set $D$.
                \EndFor
                \State merge all line segments in the set $D$ together into $\ell$.
                \State replace ${l_i^G}$ with $\ell$.
            \EndFor
        \end{algorithmic}
\end{algorithm}

After loop closing, the line segments in the global map need to be adjusted accordingly (Alg. 2). Since all original line segments are marked with the indices of corresponding robot poses, they can be easily re-transformed into the world coordinates according to the optimized robot poses (Alg. 2: lines 4-8), i.e., the original line segments in each subset ${{c_i}}$ are transformed from the local coordinates to the global coordinates. The set $D$ in Alg. 2 denotes a set that is used to store line segments which will be merged together soon. Furthermore, each line segment in the global map is updated by re-merging its corresponding original line segments, which reuses the previous recorded correspondence (Alg. 2: line 9). Since the update process for line segments in the global map is independent of each other, we use multi-threaded parallel processing mechanism to speed up map updates. Finally, the currently extracted line segments are merged with the adjusted global map using OTM-ILSM.

It should be noted that in this subsection the previous recorded correspondence between the original line segments and the corresponding final line segments is reused. This correspondence is established during the incremental line segment merging. In this process, the inter-frame error of the pose estimation provided by the scan matching module is actually small. In addition, the thresholds of the fusion conditions are also set appropriately small. Under the above conditions, it is reasonable to consider that the data association between consecutive scans is accurate. This assumption is standard and widely used in the SLAM field, and many excellent SLAM algorithms are based on such assumptions \cite{3,34}. Therefore, in this paper we assume that the correspondence established and recorded during the incremental line segment merging process remains valid after loop closing.

\section{Metrics}
In this section, we detail the newly proposed correlation-based metric for quality evaluation of line segment maps. Furthermore, a simplified error metric based on the quality metric proposed in \cite{5} is introduced. In summary, we will use the above two metrics to evaluate line segment maps in a comprehensive way.

\subsubsection{Lookup table}
\begin{figure}[t]
    \centering
    \subfigure[]{\includegraphics[scale=0.21]{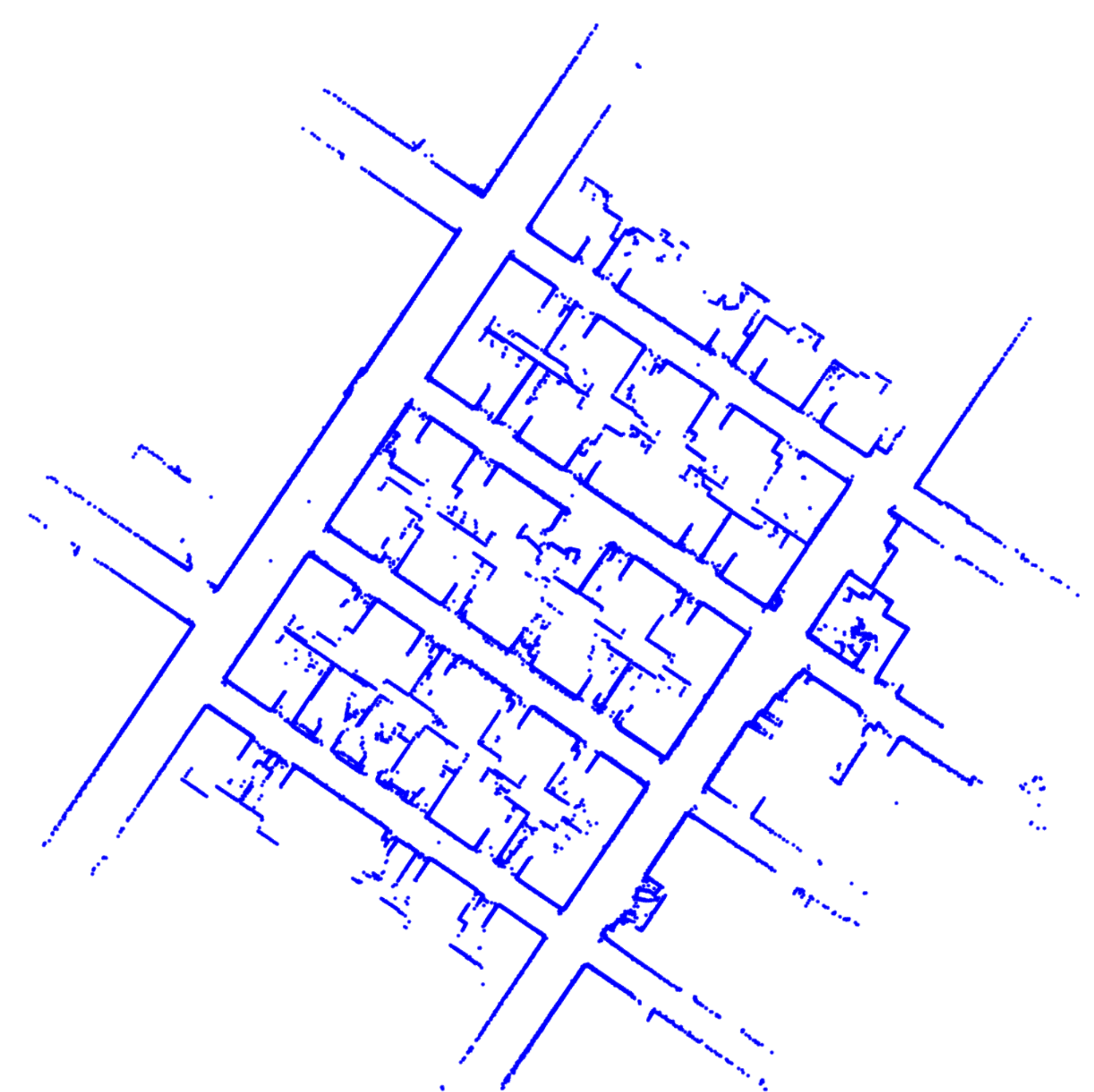}}
    \centering
    \subfigure[]{\includegraphics[scale=0.21]{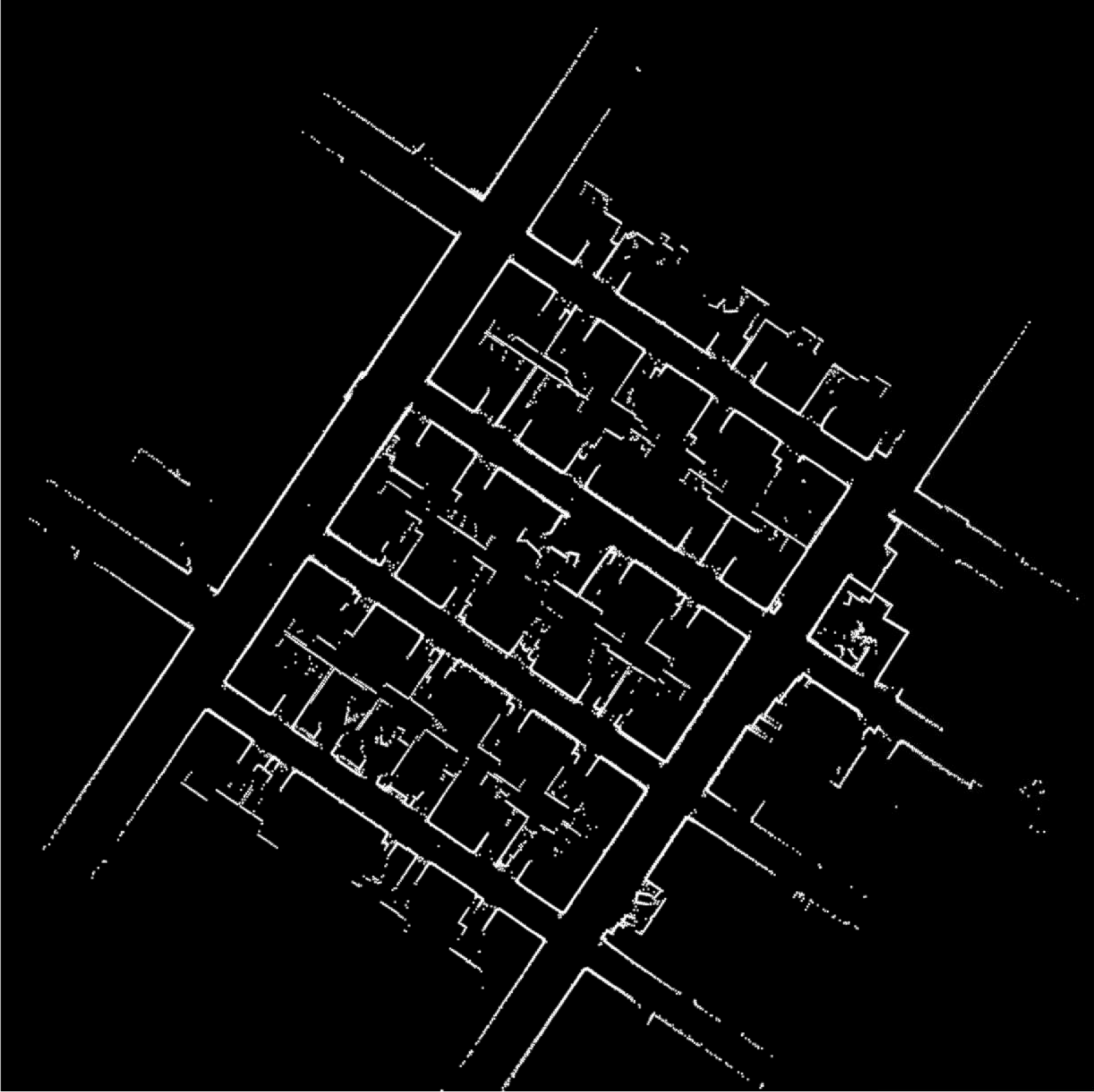}}
    \centering
    \subfigure[]{\includegraphics[scale=0.11]{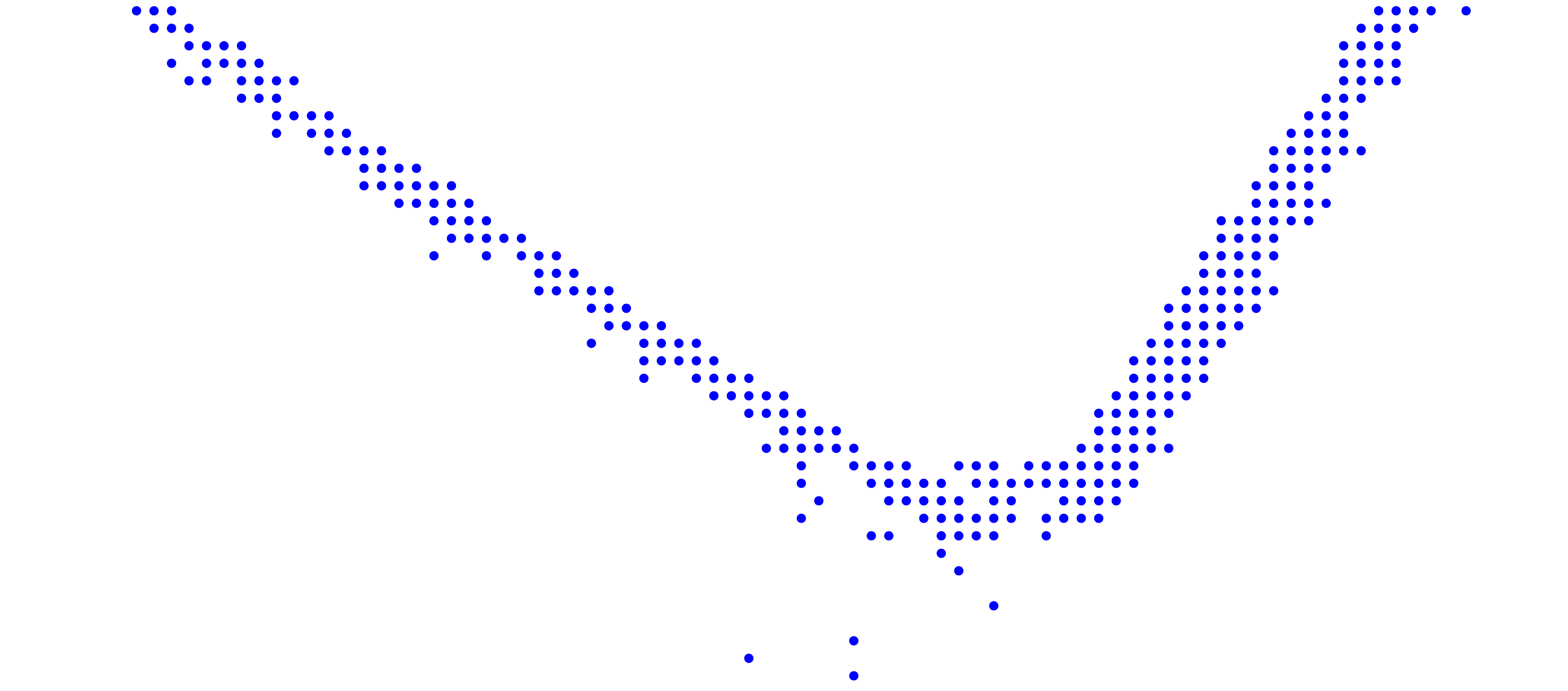}}
    \centering
    \subfigure[]{\includegraphics[scale=0.11]{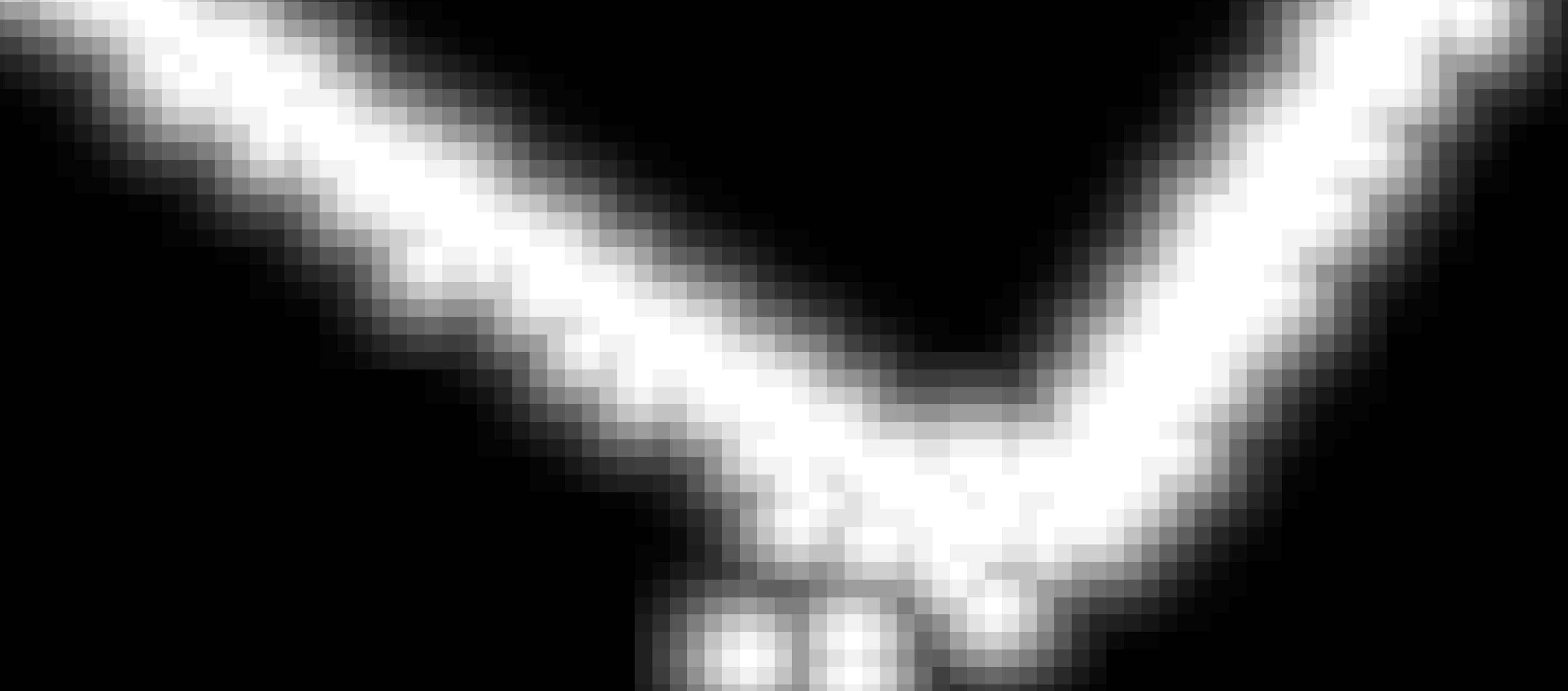}}
    \caption{(a) Registered laser scan points; (b) schematic image of lookup table; (c) local detail of registered laser scan points; (d) local detail of the corresponding lookup table. The grid cell size, i.e., the resolution of the lookup table is set to 0.01 m, and the standard deviation of the Gaussian kernel is set to 0.03 m. For each laser scan point, a circle centred on it with a radius of two standard deviations is calculated. And the grid cells covered by this circle are recorded and the occupancy probability of each covered grid cell is computed. Bright regions indicate high probability of occupancy.}
    \label{lookuptable}
\end{figure}

In order to quantitatively evaluate line segment maps, in this work we borrow the idea of correlative scan matching algorithm \cite{20} and propose a correlation-based metric. Firstly, scan matching and loop closing are used to correct robot poses. The output pose trajectory is regarded as quasi ground truth. Secondly, all laser scan points are transformed to the world coordinates according to the corresponding robot poses. Then, the laser scan points represented in continuous space are transformed to discrete space (the grid map) by a given grid cell size. Finally, a probabilistic grid map named lookup table is constructed by using a Gaussian kernel within a small neighborhood of each laser scan point. The value of each cell in the lookup table represents the probability of occupancy at that position. In this work, the grid cell size, i.e., the resolution of the lookup table is set to 0.01 m. The Gaussian kernel is used to blur laser scan points to take into account the sensor noise. Therefore, the standard deviation of the Gaussian kernel should be physically consistent with the accuracy of laser range finders. In this work, the standard deviation of the Gaussian kernel is set to 0.03 m according to the specification of the laser range finders used in the experiments. For each laser scan point, a circle centred on it with a radius of two standard deviations is calculated. And then the grid cells covered by this circle are recorded. Finally, the occupancy probability of each covered grid cell is computed. If the occupancy probability of a cell is computed multiple times, the maximum occupancy probability is taken as its cell value. The example of the lookup table is shown in Fig. \ref{lookuptable}.

Actually, the standard deviation of Gaussian kernel should not only take into account the accuracy of laser range finders but also the uncertainty associated with the robot poses. However, the uncertainty of pose estimation is usually determined by SLAM systems, while the proposed approach should be regarded as a generic add-on independent module for SLAM systems and has no effect on their performance. Therefore, in this work we suppose that the pose estimation errors are sufficiently small and do not take into account the pose estimation errors in the design and evaluation of redundant line segment merging approaches, which is also a common assumption of the related researches. For more information about the uncertainty of line segment endpoint features, please refer to our previous work on active SLAM \cite{32}.

\textsl{Remark 3:} In this work we choose two standard deviations instead of three to smear the laser scan points, which is chosen the same as the parameter settings of the correlative scan matching module \cite{20} in two open-source laser SLAM systems: SRI Karto \cite{29} and Google Cartographer \cite{3}. In addition, two standard deviations around the mean correspond to the 95\% confidence level, which is empirical enough for considering the sensor noise.

\subsubsection{Correlation-based metric}
Since the lookup table can be considered as a discrete function, we need to discretize each line segment in the final map to pixels and then project these pixels and find the corresponding scores in the lookup table. These final line segments are first scaled to the grid coordinates according to the resolution of the lookup table and then discretized to pixels by the Bresenham algorithm. Furthermore, in order to take into account the direction of line segment features, each line segment pixel is accompanied with direction information obtained by converting continuous version of the heading of its corresponding final line segment into discrete angular space. In this work, the resolution of the angular discretization is set to 1.0$^\circ$. The score of each final line segment is defined as the total scores of the cells through which the line segment passes.

\begin{figure}[t]
    \centering
    \includegraphics[scale=0.8]{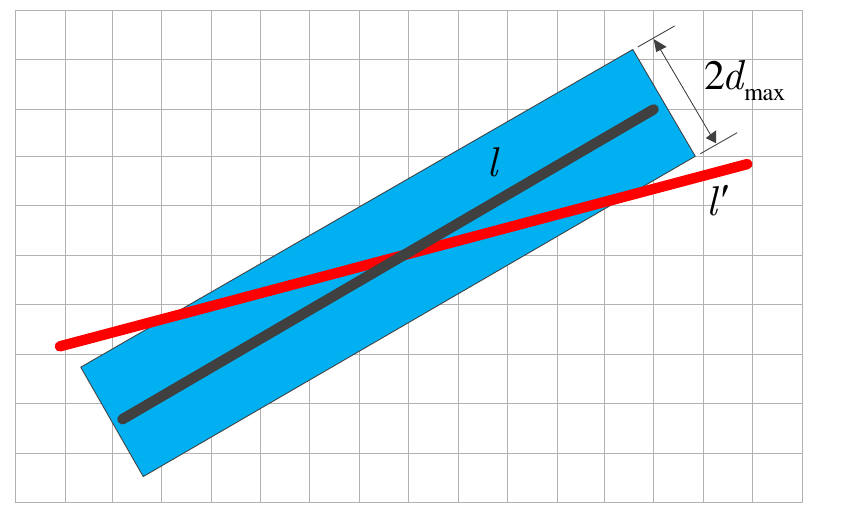}
    \caption{For one final line segment $l$, a strip with a width of $2{d_{\textup{max}}}$ is centred on it. If a certain percentage of line segment pixels from another line segment $l'$ get superimposed on the covered grid cells of the strip of $l$, these two line segments are judged as a pair of redundant line segments, and both of their scores will be used as negative items in the map quality calculation.}
    \label{overlap_metric}
\end{figure}

In addition, in order to consider the situation where there exist redundant line segments in the final map, the following approach is adopted. As shown in Fig. \ref{overlap_metric}, for each final line segment $l$, a strip with a width of $2{d_{\textup{max}}}$ is centred on it, where ${d_{\textup{max}}}$ is the threshold of the separation distance in the fusion conditions. And then the covered grid cells within this strip are recorded. The direction of these cells is consistent with the discretized angle of $l$. If a certain percentage (in this work 20\% is used) of line segment pixels from another line segment $l'$ get superimposed on the covered grid cells of the strip of $l$ and the discretized angular deviation between $l$ and $l'$ is within ${\theta _{\textup{max} }}$, where ${\theta _{\textup{max} }}$ is the threshold of the heading deviation in the fusion conditions, these two line segments are judged as a pair of redundant line segments, and both of their scores will be used as negative items in the map quality calculation. Only the scores of those non-redundant final line segments are used as positive items in the map quality calculation. Based on the above approach, the physical meaning of judging redundant line segments in continuous space and discrete space becomes consistent with each other, i.e., the judgement of redundancy by using the fusion conditions and correlation-based metric is consistent. Once there are redundant line segments in the final map, the proposed metric will penalize the map quality accordingly. It should be noted that the map quality is only related to the cells through which the line segments pass. In other words, the covered grid cells of the strip of the line segment are only calculated and stored for judgement of redundancy and not involved in map quality calculation. Finally, it should be emphasized that the final line segments are checked one by one. The line segment to be checked is only compared with those line segments which have been checked and considered as non-redundant line segments. Namely, each final line segment will be considered as a redundant line segment at most once. And a pair of redundant line segments will contribute to the penalty term only once.

The map quality of a final line segment map ${M_t} = \left\{ {l_1^G,l_2^G, \ldots ,l_k^G} \right\}$ is defined as follows:
\begin{equation}
    q = \frac{1}{N}\left( {\sum\limits_{i = 1}^{N - n} {L\left( {{p_i}} \right) - \lambda \sum\limits_{j = 1}^n {L\left( {{p_j}} \right)} } } \right),
\end{equation}
where $p$ denotes the discrete line segment pixel, $N$ is the total number of line segment pixels obtained by discretizing all final line segments $\left\{ {l_1^G,l_2^G, \ldots ,l_k^G} \right\}$, and $n$ is the number of redundant line segment pixels obtained by discretizing all redundant line segments in ${M_t}$. The map quality is defined as the average difference between the total scores of line segment pixels without redundancy and the total scores of line segment pixels with redundancy multiplied by the penalty coefficient $\lambda$. The larger the parameter $\lambda$, the greater the impact of redundant line segments on the final map quality. Normally, we choose $\lambda  \ge 1$ so that the quality measure is sufficiently sensitive to redundant line segments. In this work, $\lambda$ is set to 1, which means that redundant line segments have the same weight as non-redundant line segments. The function $\textsl{L}\left(  \bullet  \right)$ is the lookup table which takes a pixel and returns a probability of observing obstacles at that position.

\textsl{Remark 4:} In general, the number of the redundant line segments in the final map obtained by redundant line segment merging approaches is much smaller than the number of the non-redundant line segments. Therefore, the value of the map quality will not be negative in general.

The physical meaning of the proposed correlation-based metric can be summarized as follows. Firstly, the globally consistent laser scans are regarded as ground truth to describe the environment. The coincidence degree between the final line segments and the registered laser scan points is regarded as the degree of conformity between the final line segment map and the real environment. Secondly, in order to describe the environment more realistically, each laser scan point is smeared with a Gaussian kernel to take into account the sensor noise. The final lookup table can be regarded as the environmental model, which is used for the evaluation of the final line segment map. Thirdly, line segments in the final map are discretized to directed pixels and projected to match the lookup table, and the redundancy of the final map is also considered.

It should be noted that the correlation-based metric is a resolution-dependent metric. However, this evaluation metric is not a monotone function of resolution. The registered laser scan points and discrete line segment pixels will hit in different grid cells according to different resolutions. As a result, the change of the map quality with changing resolutions is not regular. In principle, this evaluation metric is meaningful only if different approaches are evaluated with the same given resolution.

\subsubsection{Error metric}
In addition to the proposed evaluation metric, we also introduce a simplified error metric based on the quality metric proposed in \cite{5}. Since each line segment in the final map has its corresponding original line segments before merging, we calculate the average distance of the center points of these original line segments to the corresponding final line segments to define the error metric:
\begin{equation}
e = {{\sum\limits_{i = 1}^K {\sum\limits_{j = 1}^{{k_i}} {\frac{{\left| {{A_i}x_j^{CG} + {B_i}y_j^{CG} + {C_i}} \right|}}{{\sqrt {A_i^2 + B_i^2} }}} } } \mathord{\left/
 {\vphantom {{\sum\limits_{i = 1}^K {\sum\limits_{j = 1}^{{k_i}} {\frac{{\left| {{A_i}x_j^{CG} + {B_i}y_j^{CG} + {C_i}} \right|}}{{\sqrt {A_i^2 + B_i^2} }}} } } {\sum\limits_{i = 1}^K {{k_i}} }}} \right.
 \kern-\nulldelimiterspace} {\sum\limits_{i = 1}^K {{k_i}} }},
\end{equation}
where $({A_i},{B_i},{C_i})$ denotes the line parameters of the final line segment, $(x_j^{CG},y_j^{CG})$ denotes the center point of the original line segment, $K$ is the number of line segments in the final map and ${k_i}$ is the number of the corresponding original line segments of each final line segment. This error metric represents the average deviation of the original line segments to the final map, which evaluates the loss of map information deriving from redundant line segment merging.

In addition, in order to compare different redundant line segment merging approaches more objectively, we also report the experimental results based on the quality metric presented in [5], which will be called \textit{distance metric} in the experiments. For the distance metric [5], the distance and angular deviations between the final line segments and the corresponding original line segments are firstly calculated. And then the weighted summation of the distance and angular deviations is defined as the loss of map information deriving from redundant line segment merging. While the proposed error metric evaluates the loss of map information through calculating the average distance of the center points of the original line segments to the corresponding final line segments. In principle, the idea of the distance metric [5] and the proposed error metric are consistent. Both of these two metrics reflect the average deviation of the original line segments and the corresponding final line segments in the geometric space.

It is worth noting that the proposed simplified error metric only makes sense when the sensor and pose estimation errors are zero or sufficiently small since we do not take into account the sensor and pose estimation errors when designing this metric. Actually, the sensor and pose estimation errors can be considered as system noise, which are usually considered in the SLAM algorithms. While as mentioned before, the proposed approach should be regarded as an add-on module of SLAM systems which supposes that the pose estimation errors are sufficiently small. In the following comparative experiments, we will use the same line segment extraction algorithm and pose estimation approach. Based on the above process, the sensor and pose estimation errors will have the same effect on all redundant line segment merging approaches involved in the comparative experiments. Therefore, the objectivity and fairness of the comparative experiments based on the simplified error metric can be guaranteed.

\subsubsection{Discussion}
The correlation-based metric compares the final line segment map with the registered laser scan points to evaluate the coincidence degree between the final line segment map and the environment. And the error metric evaluates the average deviation between the final line segments after merging and the corresponding original line segments before merging. Therefore, the correlation-based metric and error metric are essentially two different types of evaluation metrics. These two types of metrics are proposed and used to evaluate the performance of different redundant line segment merging approaches in a comprehensive manner. In addition, in order to demonstrate the validity of the error metric and its consistency with the distance metric, the evaluation results based on the error metric and distance metric are reported in the paper.

\section{Experiment}
In this section, the experimental setup and results of the comparative experiments with respect to MS-RLSM \cite{6} and OTO-ILSM \cite{19} are detailed. We first introduce the public datasets and self-recorded dataset used in the experiments. And then the comparative experimental results and qualitative and quantitative analysis are presented to show the superior performance of CAE-RLSM in terms of efficiency and map quality in different scenarios.

Before introducing the datasets used in the experiments, we first introduce the general idea of O$^{2}$TO-ILSM designed and used in the comparative experiments. To overcome the issues of online OTO-ILSM as mentioned before, we design an offline version of OTO-ILSM for global map adjustment after loop closing, called O$^{2}$TO-ILSM. Firstly, the open-source 2D laser SLAM system Karto\footnote{\textcolor[rgb]{0.00,0.00,1.00}{\url{https://github.com/ros-perception/open_karto}}} is used to obtain the globally consistent laser scans, and the original line segments extracted from raw laser scans are orderly recorded according to the order of laser scans. These line segments are represented in the world coordinates according to the optimized poses. Secondly, the original line segments are processed by O$^{2}$TO-ILSM according to the order of laser scans. Based on the above scheme, O$^{2}$TO-ILSM and CAE-RLSM can be compared fairly after loop closing.

\begin{table*}[t]
    \centering
    \caption{Quantitative statistics of the raw datasets and the original line segments before merging}
    \label{table_1}
    \scalebox{0.9}{
    \begin{tabular}{cccccccc}
        \toprule
        \multicolumn{1}{c}{}        & \# Scan proc.     & Avg. $\Delta t$   & Avg. $\Delta \theta$  & \# Scan line      & Shortest line     & Longest line      & Average length        \\
        \multicolumn{1}{c}{}        &                   & (mm)              &                       & segments          & segment (mm)      & segment (mm)      &  (mm)                 \\
        \midrule
        \bm{$[1]$}                  &  \bm{$[2]$}       &  \bm{$[3]$}       & \bm{$[4]$}            & \bm{$[5]$}        & \bm{$[6]$}        & \bm{$[7]$}        & \bm{$[8]$}            \\
        \midrule
        Dataset (a)                 & 1413              & 176.8             &  4.40$^\circ$         & 4432              &  600.0            & 11447.0           & 2129.8                \\
        Dataset (b)                 & 1192              & 197.2             &  4.16$^\circ$         & 3649              &  600.0            &  7937.5           & 1767.2                \\
        Dataset (c)                 &  240              & 794.8             & 11.94$^\circ$         & 1167              &  600.0            &  5793.0           & 1700.8                \\
        Dataset (d)                 &  786              & 216.8             &  1.02$^\circ$         & 4058              & 1000.7            & 11809.8           & 3113.3                \\
        \bottomrule
    \end{tabular}}
\end{table*}

\begin{table*}[!htb]
    \centering
    \caption{Quantitative statistics of partial line segment maps produced by MS-RLSM with default and optimized parameters on dataset (c)}
    \label{table_2}
    \scalebox{0.9}{
    \begin{tabular}{cccccccc}
        \toprule
        \multicolumn{1}{c}{}    & \# Map line       & Shortest line     & Longest line  & Average length    & Quality     \\
        \multicolumn{1}{c}{}    & segments          & segment (mm)      & segment (mm)  &  (mm)             &             \\
        \midrule
        \bm{$[1]$}              &  \bm{$[2]$}       &  \bm{$[3]$}       &  \bm{$[4]$}   & \bm{$[5]$}        &  \bm{$[6]$} \\
        \midrule
        Default parameters      & 15                &  751.4            & 11761.6       & 5497.3            & 66.94\%     \\
        Optimized parameters    & 39                &  669.1            &  8959.8       & 2066.9            & 96.51\%     \\
        \bottomrule
    \end{tabular}}
\end{table*}

\begin{table*}[!htb]
    \centering
    \caption{Quantitative statistics of line segment maps produced by MS-RLSM, O$^{2}$TO-ILSM and CAE-RLSM}
    \label{table_3}
    \scalebox{0.9}{
    \begin{tabular}{ccccccccccc}
        \toprule
        \multicolumn{2}{c}{}                               & \# Map line  & Shortest line  & Longest line  & Average      & Quality  & Error  & Distance   & Per-frame     & Total offline      \\
        \multicolumn{2}{c}{}                               & segments     & segment (mm)   & segment (mm)  & length (mm)  &          & (mm)   & (mm)       & runtime (ms)  & runtime (ms)       \\
        \midrule
        \bm{$[1]$}  & \bm{$[2]$}    & \bm{$[3]$}    & \bm{$[4]$}    & \bm{$[5]$}    & \bm{$[6]$}    & \bm{$[7]$}   & \bm{$[8]$}  & \bm{$[9]$} & \bm{$[10]$} & \bm{$[11]$} \\
        \midrule
        \multirow{2}{*}{Dataset (a)}  &   MS-RLSM          & 125        & 671.9          & 19668.3       & 2945.9       & 92.83\%       & 10.25       & 19.25      & N/A      & 8440.7    \\
                                      &   O$^{2}$TO-ILSM   & 130        & 671.2          & 19593.3       & 2832.8       & 90.80\%       & 10.44       & 18.16      & 0.061    & 72.0      \\
                                      &   CAE-RLSM         & 126        & 672.1          & 19724.6       & 2923.1       & \bf{94.88}\%  &  \bf{8.34}  & \bf{16.92} & 0.076    & 75.1      \\
        \midrule
        \multirow{2}{*}{Dataset (b)}  &   MS-RLSM          & 128        & 753.9          & 14692.6       & 2138.7       & 94.87\%       &  7.62       & 10.61      & N/A      & 5343.1    \\
                                      &   O$^{2}$TO-ILSM   & 131        & 741.6          & 14695.9       & 2081.8       & 90.66\%       &  8.06       & 9.89       & 0.055    & 55.1      \\
                                      &   CAE-RLSM         & 130        & 718.7          & 14717.5       & 2134.0       & \bf{95.75}\%  &  \bf{7.49}  & \bf{9.32}  & 0.082    & 77.2      \\
        \midrule
        \multirow{2}{*}{Dataset (c)}  &   MS-RLSM          & 94         & 669.1          & 8959.8        & 1916.5       & 95.80\%       & 11.17       & 9.13       & N/A      & 730.3     \\
                                      &   O$^{2}$TO-ILSM   & 96         & 669.5          & 8958.8        & 1869.6       & 93.32\%       &  4.69       & 5.34       & 0.076    & 18.2      \\
                                      &   CAE-RLSM         & 93         & 668.5          & 8960.8        & 1909.4       & \bf{97.68}\%  &  \bf{3.81}  & \bf{5.10}  & 0.083    & 25.4      \\
        \midrule
        \multirow{2}{*}{Dataset (d)}  &   MS-RLSM          & 109        & 1094.5         & 13603.6       & 4120.8       & 97.73\%       &  9.04       & 8.70       & N/A      & 5632.7    \\
                                      &   O$^{2}$TO-ILSM   & 114        & 1101.3         & 13568.8       & 4051.5       & 90.31\%       & 13.70       & 10.16      & 0.070    & 64.2      \\
                                      &   CAE-RLSM         & 110        & 1095.5         & 13604.6       & 4088.3       & \bf{97.87}\%  &  \bf{8.76}  & \bf{8.46}  & 0.082    & 67.9      \\
        \bottomrule
    \end{tabular}}
\end{table*}

\subsection{Datasets}
The following three public datasets obtained from the Robotics Data Set Repository (Radish) \cite{15} and one dataset recorded in our lab are considered in the comparative experiments:

\begin{enumerate}[(a)]
\item department\_diiga: this dataset is recorded in the Department of DIIGA at Engineering University in Ancona. The dataset is collected by using a mobile robot equipped with a SICK LMS 200 laser range finder with 1.0$^\circ$ angular resolution. The area of the environment is approximately $47m$ \( \times \) $47m$ and the number of scans is 8540.
\item intel\_oregon: this dataset is recorded in the part of the Intel Lab in Hillsboro, Oregon. The dataset is collected by using a mobile robot equipped with a SICK LMS 200 laser range finder with 1.0$^\circ$ angular resolution. The area of the environment is approximately $23m$ \( \times \) $23m$ and the number of scans is 8030.
\item seattle: this dataset is relative to University of Washington in Seattle, Washington. The dataset is collected by using a mobile robot equipped with a SICK LMS 200 laser range finder with 0.5$^\circ$ angular resolution. The area of the environment is approximately $52m$ \( \times \) $13m$ and the number of scans is 241.
\item floor\_4: this dataset is collected in a long hallway with a perfect loop in our laboratory by using a mobile robot equipped with a SICK LMS 100 laser range finder with 0.25$^\circ$ angular resolution. The area of the environment is approximately $75m$ \( \times \) $55m$ and the number of scans is 20384.
\end{enumerate}
The original laser scans on dataset (c) have been already pre-registered, so the tests on this dataset do not require pose correction, which demonstrates that CAE-RLSM is independent of loop closure detection and pose optimization.

\begin{figure*}[htb]
    \centering
    \subfigure[]{\includegraphics[scale=0.24]{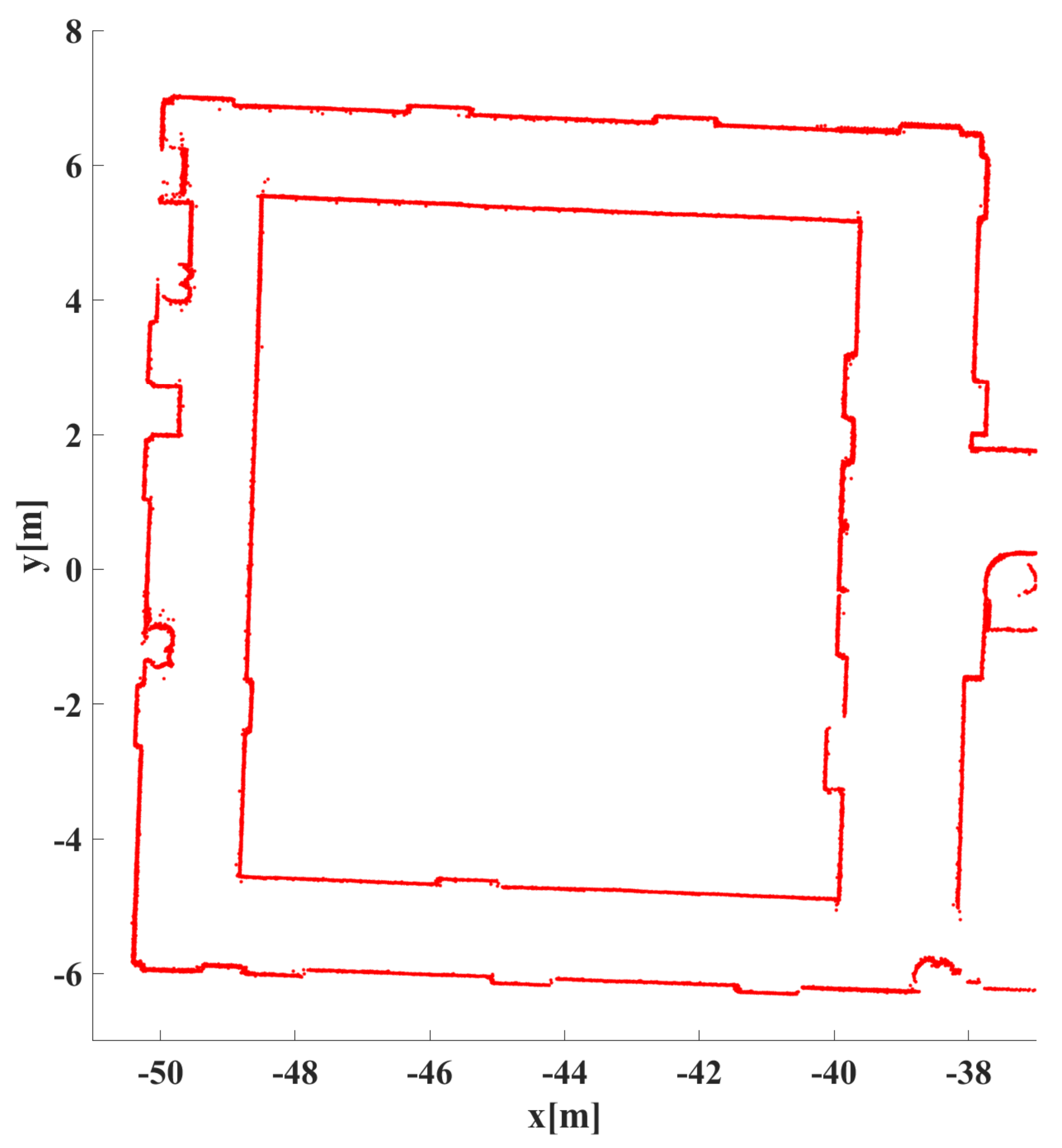}}
    \hspace{0.5 cm}
    \centering
    \subfigure[]{\includegraphics[scale=0.24]{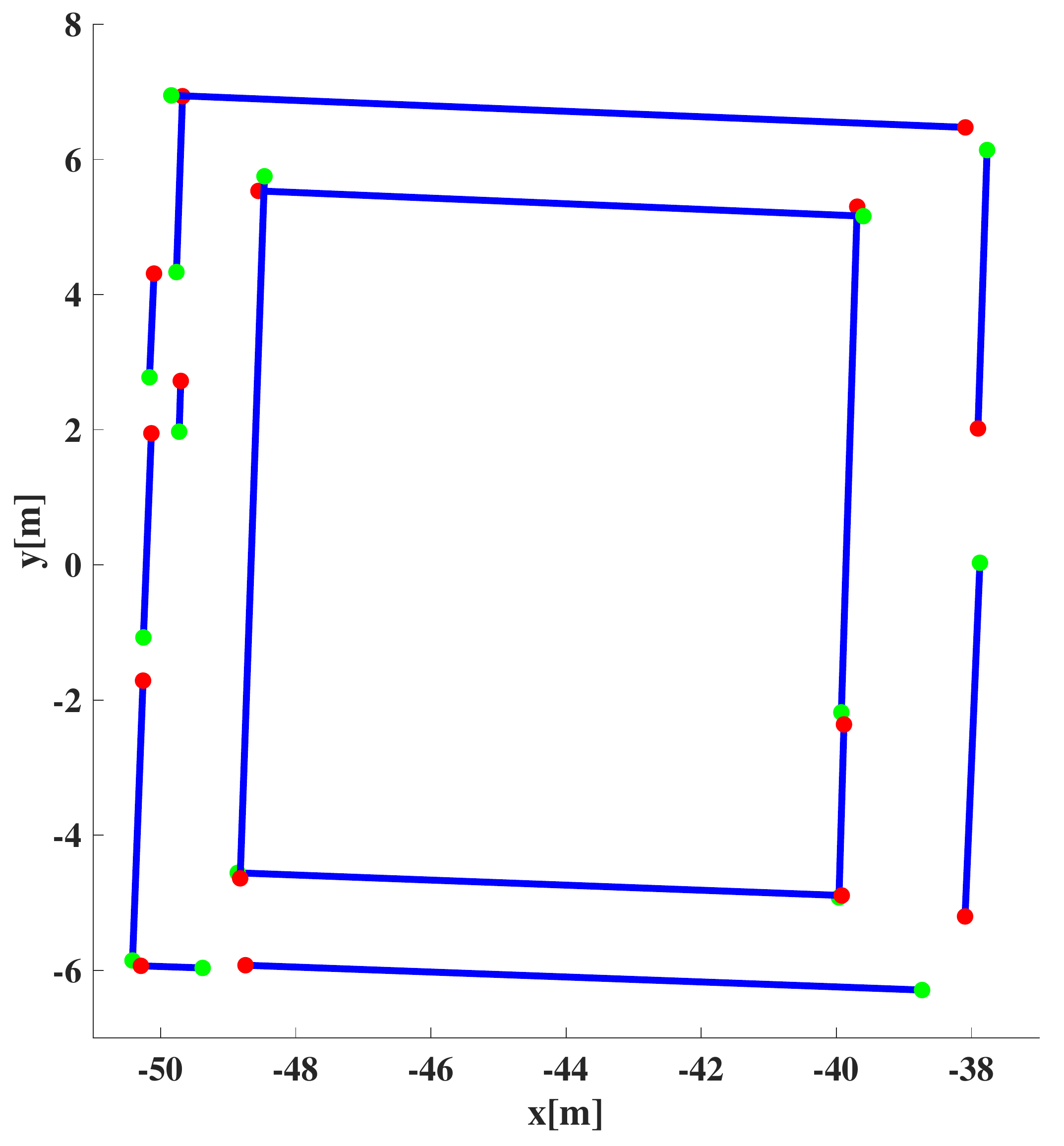}}
    \hspace{0.5 cm}
    \centering
    \subfigure[]{\includegraphics[scale=0.24]{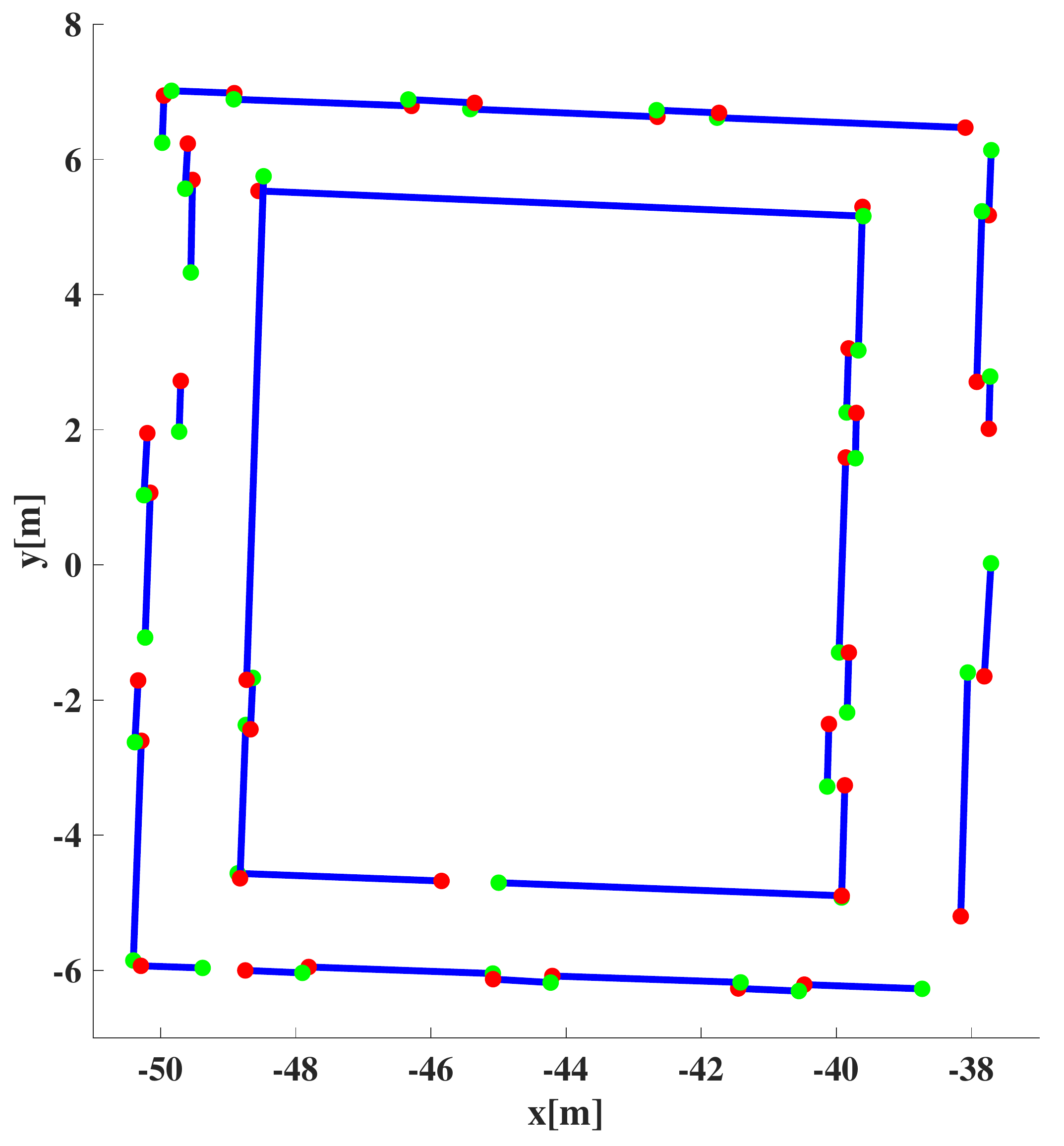}}
    \caption{Partial experimental results on dataset (c). (a) Registered laser scan points; (b) partial line segment map produced by MS-RLSM with default parameters; (c) partial line segment map produced by MS-RLSM with optimized parameters.}
    \label{seattle_part}
\end{figure*}

\begin{figure*}[t]
    \centering
    \subfigure[]{\includegraphics[scale=0.33]{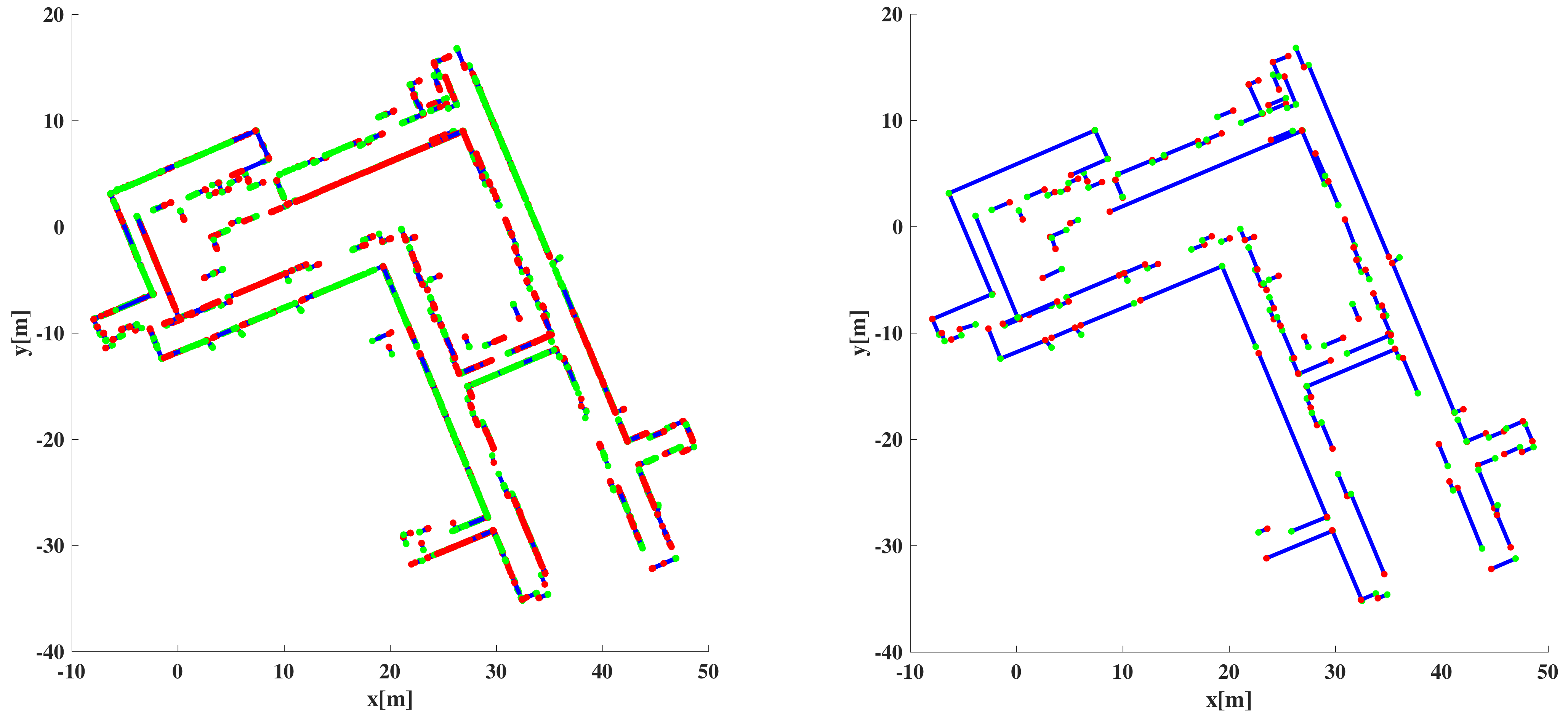}}
    \centering
    \subfigure[]{\includegraphics[scale=0.33]{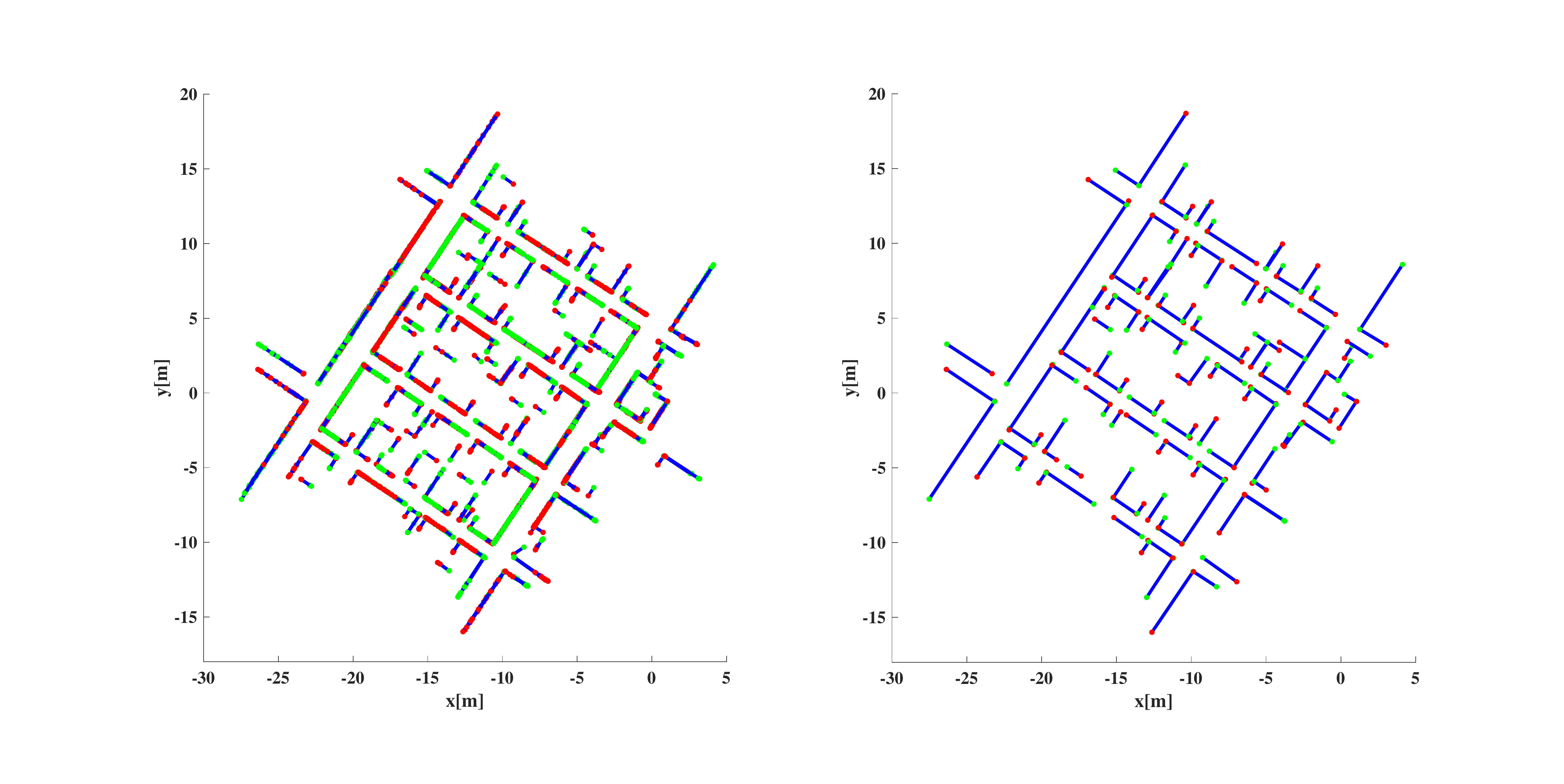}}
    \caption{Line segment maps obtained by piling up all original line segments extracted from raw laser scans (left) and final maps produced by CAE-RLSM (right) on datasets (a) and (b). The red and green dots represent the start and end points of line segments, respectively. These images are all vector graphics. Readers are advised to refer to the electronic version of this article to see the endpoints of line segments more clearly by zooming in on these images.}
    \label{final_map}
\end{figure*}

\begin{figure*}[t]
    \centering
    \includegraphics[scale=0.18]{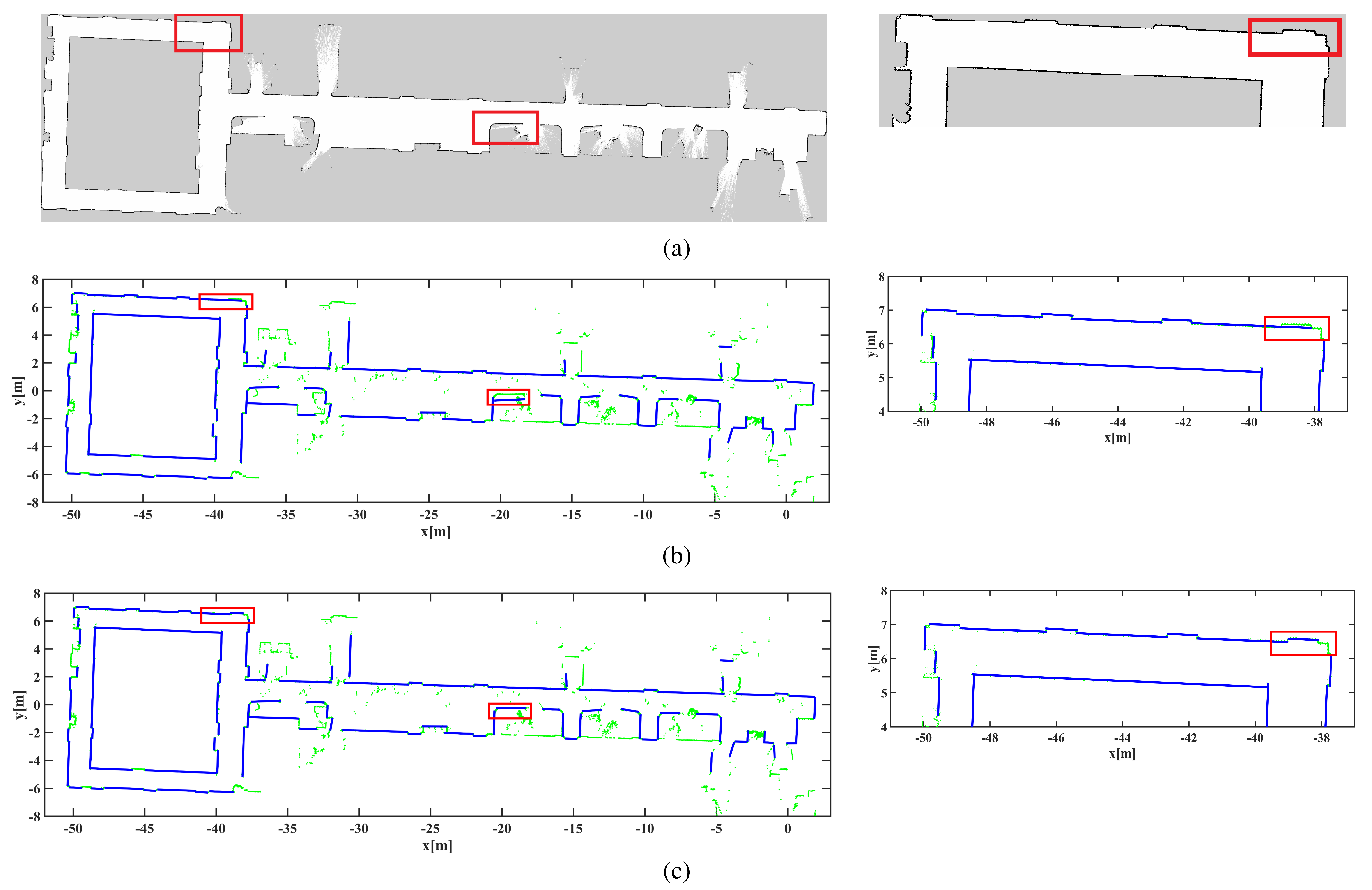}
    \caption{Experimental results on dataset (c). (a) Grid map and local detail obtained by Karto; (b) final line segment map and local detail obtained by MS-RLSM; (c) final line segment map and local detail obtained by CAE-RLSM. The green dots and blue line segments represent the registered laser scan points and line segments in the final maps, respectively. The comparative details of the coincidence degree between the final maps and the registered laser scans are circled with red rectangle borders, which demonstrates that CAE-RLSM can achieve superior performance in terms of map quality.}
    \label{seattle}
\end{figure*}

In this paper, we broadly follow the open-source line segment extraction algorithm\footnote{\textcolor[rgb]{0.00,0.00,1.00}{\url{https://github.com/ghm0819/laser-line-segment}}} \cite{18}, which borrows the idea of seed region growing in the field of image processing. This algorithm has been verified by experiments to achieve better performance than \emph{Iterative End Point Fit} algorithm \cite{9} in terms of efficiency, correctness, and precision. For datasets (a), (b) and (c), those line segments that are over 0.6 m and contain at least 10 laser scan points are preserved. For dataset (d), since the environment is a long hallway, we only preserve those line segments that are over 1.0 m and contain at least 15 laser scan points. Both the minimum length of line segments and the minimum number of contained laser scan points are empirical parameters. If these two thresholds are set too large, it may result in the failure to extract line segment features from some linear structures. Conversely, if these two thresholds are set too small, short line segments extracted from some irregular areas of the environment may also participate in the final map building, reducing the quality of the map. It should also be noted that the proposed redundant line segment merging approach is highly modularized so that the line segment extraction module used in this paper can be replaced by other alternatives \cite{9,27}.

Furthermore, we use the open-source 2D laser SLAM system Karto to complete the task of pose correction for both redundant line segment merging and lookup table construction. Whenever the robot moves more that 0.2 m or rotates more than 10$^\circ$, the new incoming laser scan is processed. It should be emphasized again that using the same line segment extraction algorithm and pose correction approach makes the comparison between different redundant line segment merging approaches more objective and fair.

Table \ref{table_1} enumerates a few quantitative statistics of the above four datasets. The \# Scan proc. column indicates the numbers of scans which have been processed for line segment extraction. The average translation and rotation between two consecutive laser scans are indicated in the Avg. $\Delta t$ (mm) column and Avg. $\Delta \theta$ column, respectively. The \# Scan line segment column indicates the number of original line segments extracted from raw laser scans, and the next three columns indicate the minimum length, maximum length and average length of these original line segments, respectively.

\subsection{Experimental Setup}
MS-RLSM, OTO-ILSM and CAE-RLSM are programmed in C/C++ and tested on a laptop with Intel Core i5-3230M CPU and 4 GB RAM.

In this paper, we do not use the default parameters of MS-RLSM in its original paper, i.e., the lateral separation is set to 400 mm and the longitudinal overlap is set to -100 mm, since the performance with these default parameters on the tested datasets is very poor. We set the optimized parameters for MS-RLSM, OTO-ILSM and CAE-RLSM as follows through extensive experimental tests to achieve the best performance for all of them. For datasets (a), (b) and (d), the lateral separation of MS-RLSM and the separation distance of CAE-RLSM are set to 100 mm, and the longitudinal overlap of MS-RLSM and the overlap of CAE-RLSM are set to -100 mm. The heading deviation of CAE-RLSM is set to 4$^\circ$. For dataset (c), the lateral separation of MS-RLSM and the separation distance of CAE-RLSM are set to 50 mm, and the longitudinal overlap of MS-RLSM and the overlap of CAE-RLSM are set to -50 mm. The heading deviation of CAE-RLSM is set to 2$^\circ$. The parameter settings of OTO-ILSM are consistent with CAE-RLSM.

In addition, as mentioned before, those line segments observed and updated many times are more stable and reliable, which are more valuable to be preserved. Therefore, for datasets (a), (b) and (d), we only preserve those line segments in the final maps updated more than 5 times. For dataset (c), those line segments in the final maps updated more than 3 times will be preserved since the entire dataset is sparse and it contains only 240 laser scans.

\subsection{Comparative Evaluation}

\subsubsection{Validation of the proposed metric}
First of all, the partial line segment maps produced by MS-RLSM with default and optimized parameters on dataset (c) are used to demonstrate the rationality of the proposed evaluation metric, as shown in Fig. \ref{seattle_part}. Table \ref{table_2} enumerates a few attributes of these line segment maps. In this example, if we use the number or the length of line segments in the final map to evaluate the quality of line segment maps \cite{5,6}, we will get the conclusion that the quality of the map produced by MS-RLSM with default parameters is better than that of using optimized parameters. But intuitively, the map produced by using optimized parameters is closer to the real environment. The reason for the wrong conclusion is that the rationality of the evaluation metrics based on the number or the length of line segments in the final map are based on the premise that the redundant line segments are correctly merged. Once physically different line segments are merged together by mistake, using these metrics will lead to wrong evaluation conclusions. On the contrary, the proposed correlation-based metric essentially evaluates the conformity of the line segment map to the real environment, as shown in Fig. \ref{seattle_part} and Table \ref{table_2}. Therefore, evaluation results based on the proposed metric are consistent with the visual inspection, which demonstrates the effectiveness of the proposed metric.

\begin{figure*}[t]
    \centering
    \subfigure[]{\includegraphics[scale=0.27]{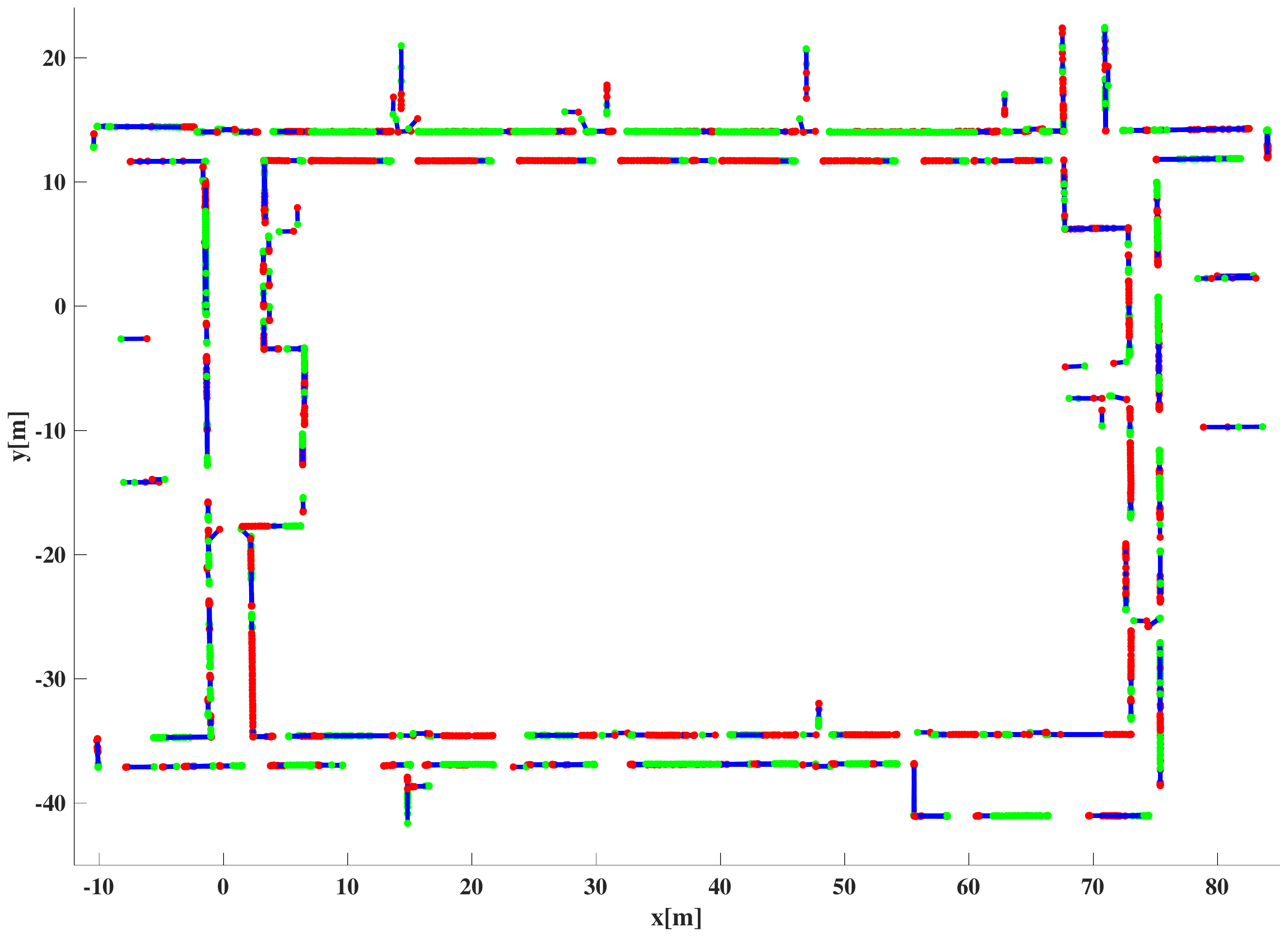}}
    \centering
    \subfigure[]{\includegraphics[scale=0.27]{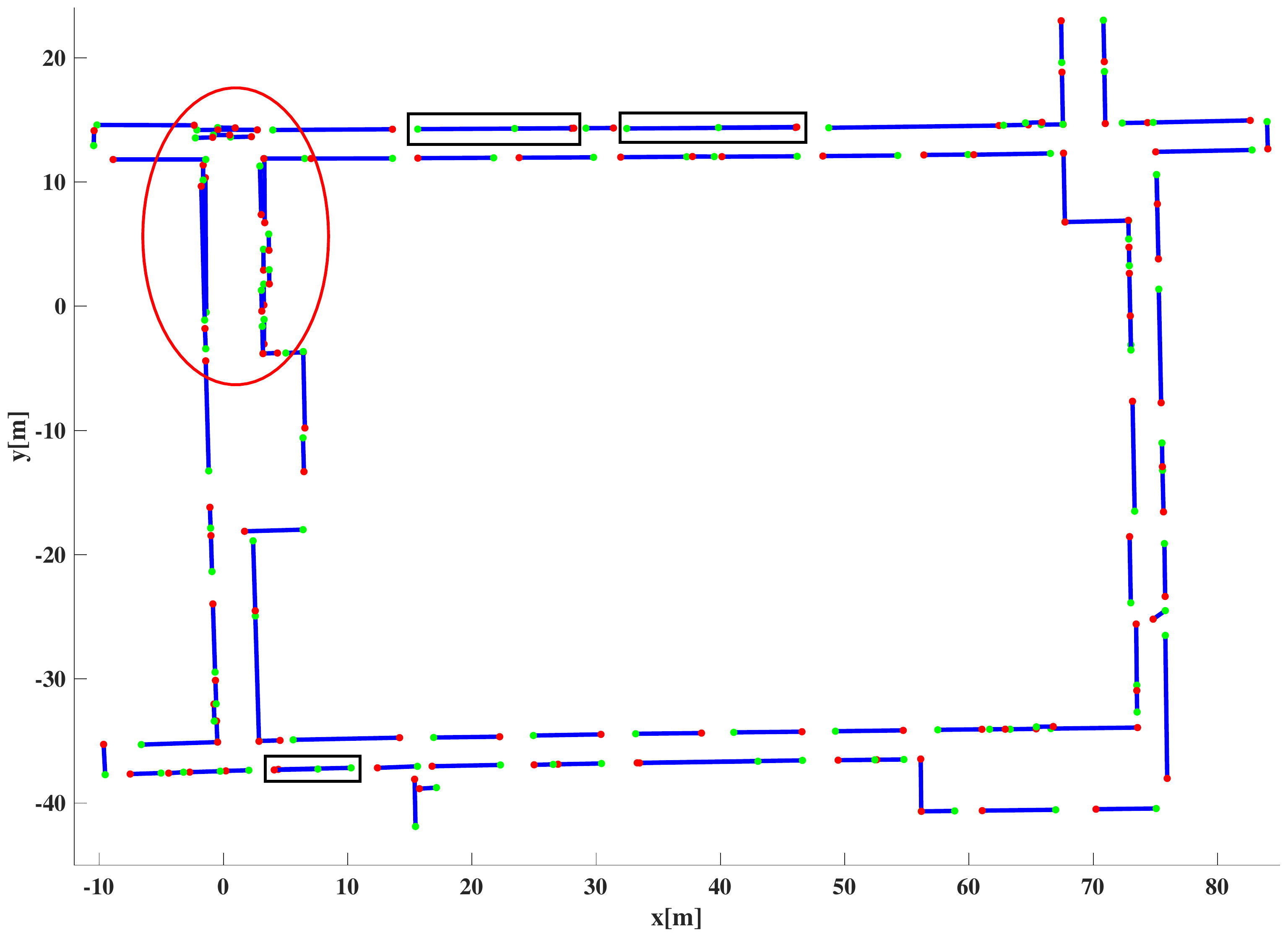}}
    \centering
    \subfigure[]{\includegraphics[scale=0.27]{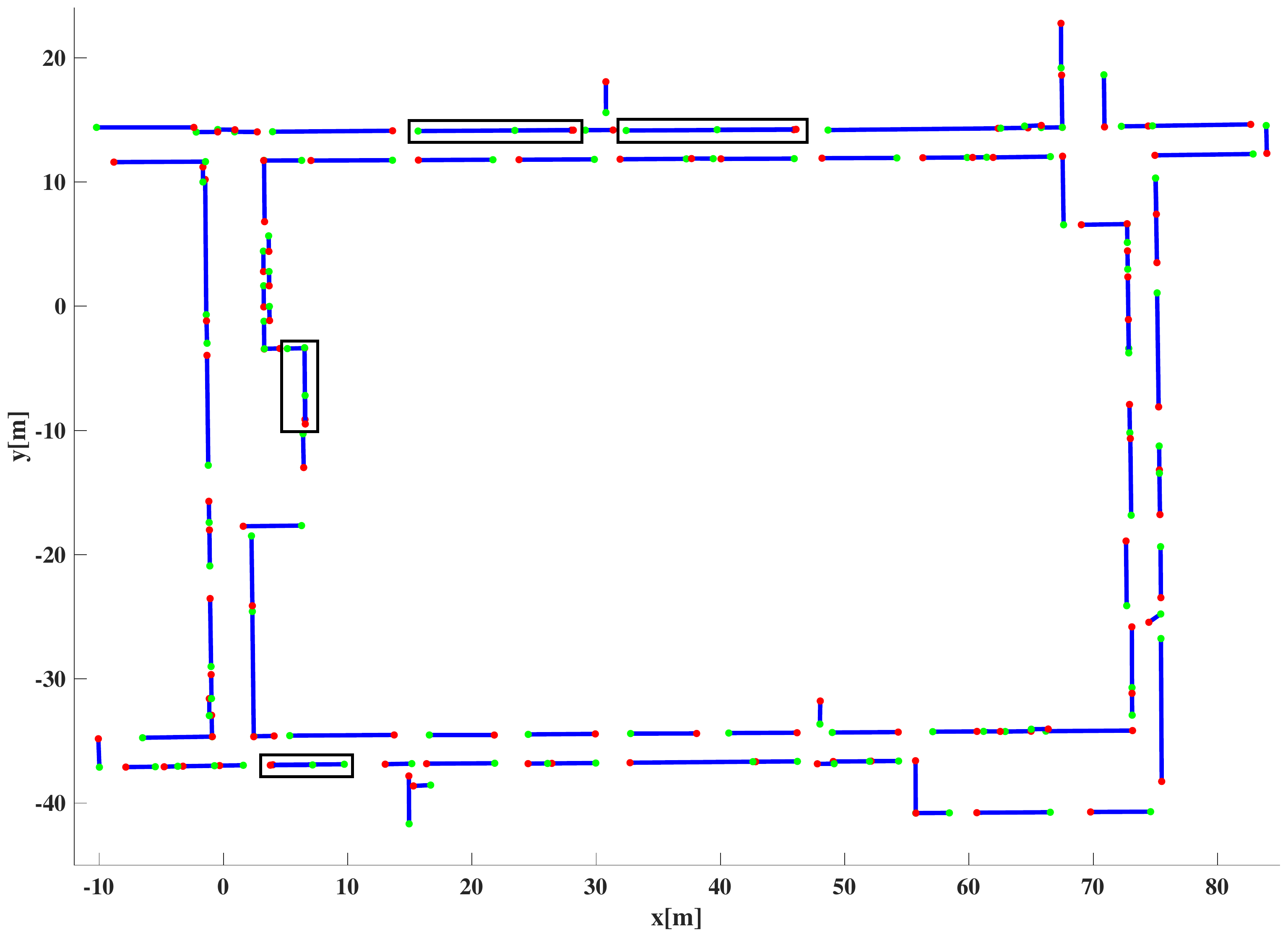}}
    \centering
    \subfigure[]{\includegraphics[scale=0.27]{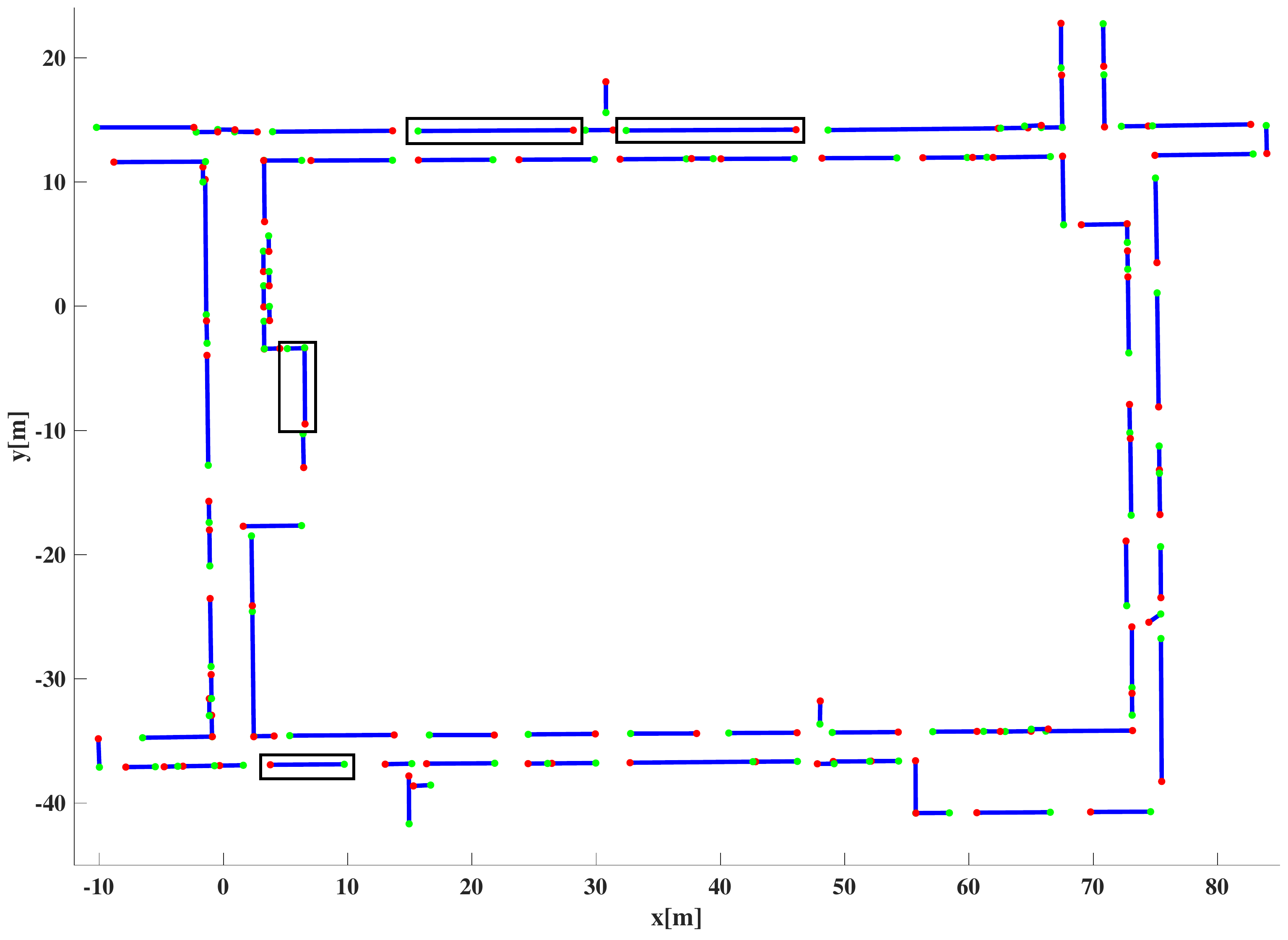}}
    \caption{Experimental results on dataset (d). (a) Original line segment map obtained by piling up all original line segments extracted from raw laser scans; (b) final map produced by online OTO-ILSM; (c) final map produced by O$^{2}$TO-ILSM; (d) final map produced by CAE-RLSM. The red and green dots represent the start and end points of line segments, respectively. The redundancy of the final map produced by OTO-ILSM is circled with black rectangle borders, while using CAE-RLSM can obtain a non-redundant line segment map. These images are all vector graphics. Readers are advised to refer to the electronic version of this article to see the endpoints of line segments more clearly by zooming in on these images.}
    \label{floor_4}
\end{figure*}

\subsubsection{Validation of the proposed approach}
Final line segment maps produced by CAE-RLSM in four datasets are shown in Figs. \ref{final_map}, \ref{seattle} and \ref{floor_4}. To the best of our knowledge, it is the first time to clearly provide the information of line segments and their endpoints in such an intuitive way. Table \ref{table_3} enumerates a few attributes of these line segment maps. The Quality column indicates the quality of line segment maps based on the correlation-based metric, the Error (mm) column and Distance (mm) column indicate the map quality based on the error metric and distance metric, respectively. It can be seen that the quality of the maps produced by CAE-RLSM is all above 94\%. In particular, the quality is above 97\% for datasets (c) and (d). On the other hand, the errors, i.e., the average deviations of original line segments to final maps introduced by CAE-RLSM are all below 10 mm, which demonstrates that CAE-RLSM can achieve good performance in terms of map quality.

\subsubsection{Map quality comparison}
In this subsection, we present the quantitative and qualitative comparison on map quality between MS-RLSM, OTO-ILSM, and CAE-RLSM.

\paragraph{MS-RLSM vs. CAE-RLSM}
Table \ref{table_3} enumerates a few attributes of the final line segment maps produced by MS-RLSM. Compared with MS-RLSM, the map quality of CAE-RLSM increases by an average of 1.4\%, the error reduces by 2.42 mm on average and the distance deviation reduces by an average of 1.97 mm. Even though the experimental results do not show a strong quantitative advantage for the proposed CAE-RLSM approach, it needs to be explained that some qualitative improvements can be thoroughly reflected in local details as shown in Fig. \ref{seattle}. To qualitatively compare these two approaches in an intuitive manner, we show the grid map generated by Karto in Fig. \ref{seattle}(a), and we also overlap the line segment maps produced by MS-RLSM and CAE-RLSM with the registered laser scan points in Fig. \ref{seattle}(b) and Fig. \ref{seattle}(c), respectively. Compared with MS-RLSM, the final map produced by CAE-RLSM is closer to the real environment. In particular, as shown in the local details of Fig. \ref{seattle}(b) and Fig. \ref{seattle}(c), MS-RLSM improperly merges two different line segments together, while CAE-RLSM preserves the structural information of the environment well. Such a specific detail can be reflected in the map quality, which demonstrates the effectiveness of the proposed evaluation metric. Furthermore, these improvements are important to facilitate the higher-level tasks such as scene recognition, accurate and robust robot localization, and so on. For example, when the line segment-based feature map in Fig. \ref{seattle}(b) is used for scene recognition, it is possible that the door in the red rectangle is mistakenly lost; when this map is used for robot localization, the online feature association is possible to be wrong, so that the localization error of the robot could increase to more than half a meter. Such robustness issue needs to be considered in the field of robot perception and SLAM \cite{1}.

\paragraph{OTO-ILSM vs. CAE-RLSM}
It should be noted that OTO-ILSM can be run in an online manner frame by frame. However, in this case, this approach is not able to adjust the merged line segments in the stage of loop closing, and thus the online OTO-ILSM presents poor performance, as shown in the area circled with the red elliptic ring in Fig. \ref{floor_4}(b). Different from OTO-ILSM, the proposed CAE-RLSM records all the line segments in different frames and maintains a subset for each physically same line segment, and thus the global map adjustment can be conducted by adjusting all the elements in the same subset according to the optimized pose graph followed by a re-merging process. Therefore, it is seen that the proposed CAE-RLSM presents superior performance than online OTO-ILSM.

\begin{figure}[t]
    \centering
    \subfigure[]{\includegraphics[scale=0.09]{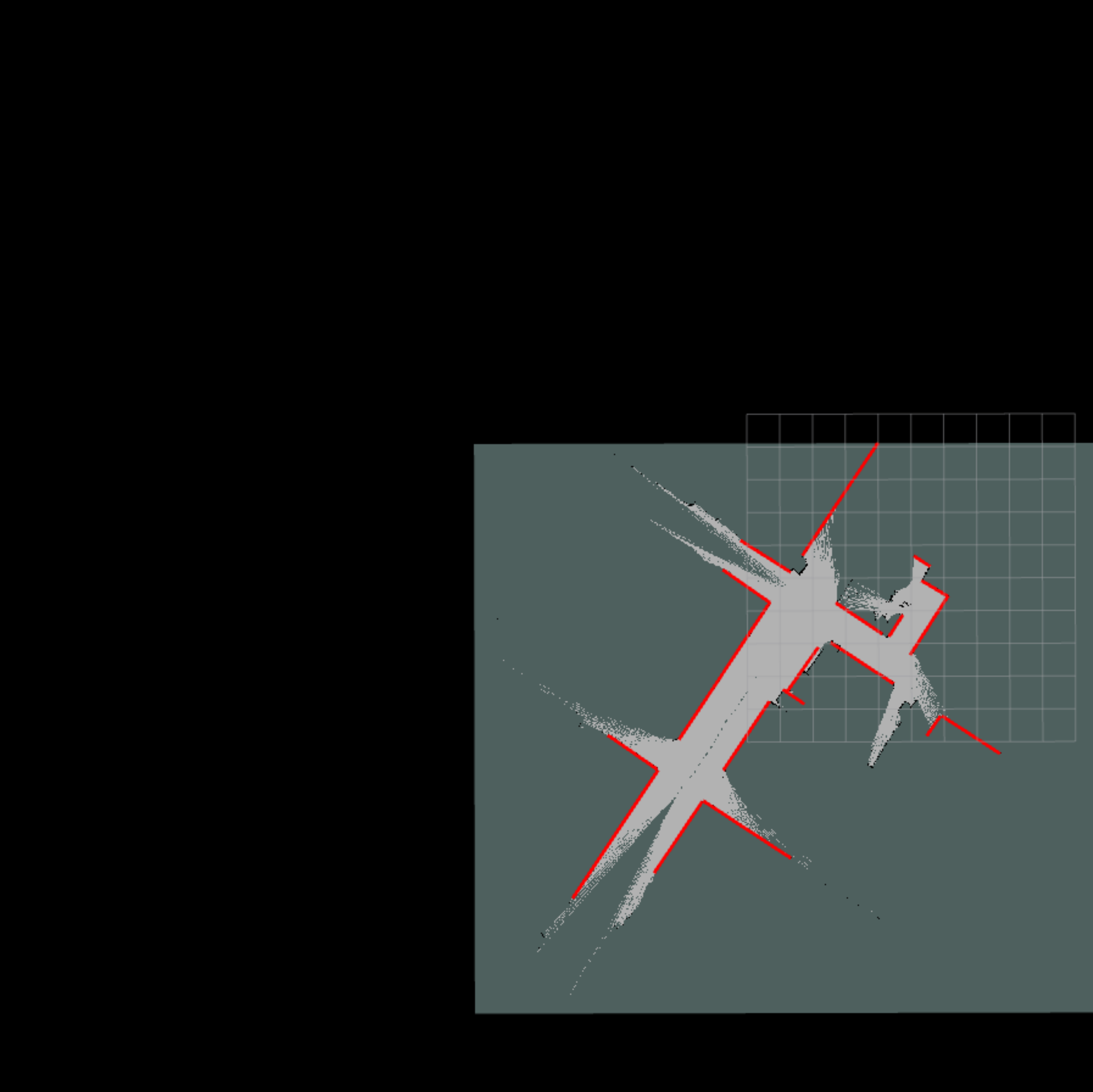}}
    \centering
    \subfigure[]{\includegraphics[scale=0.09]{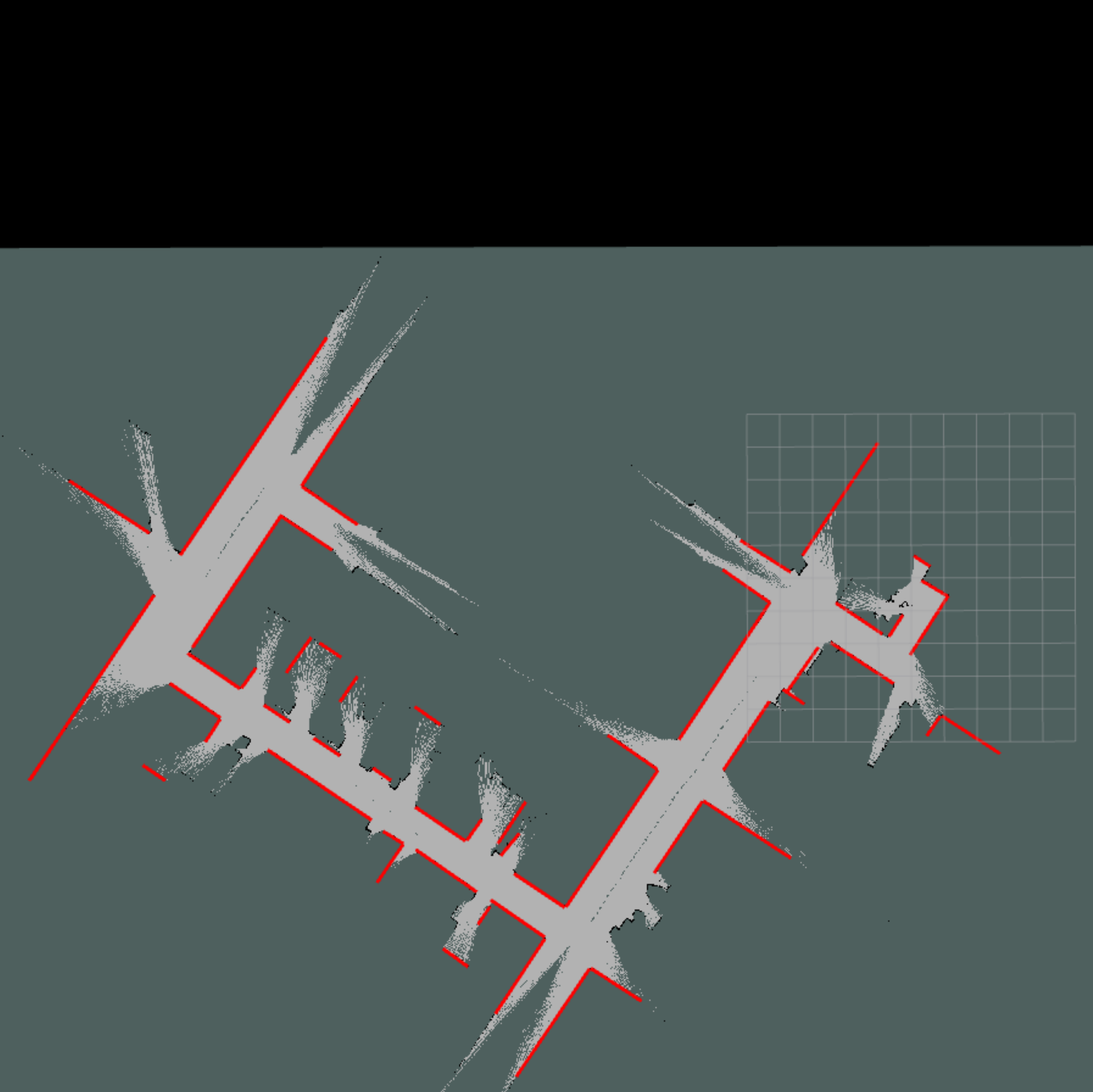}}
    \centering
    \subfigure[]{\includegraphics[scale=0.09]{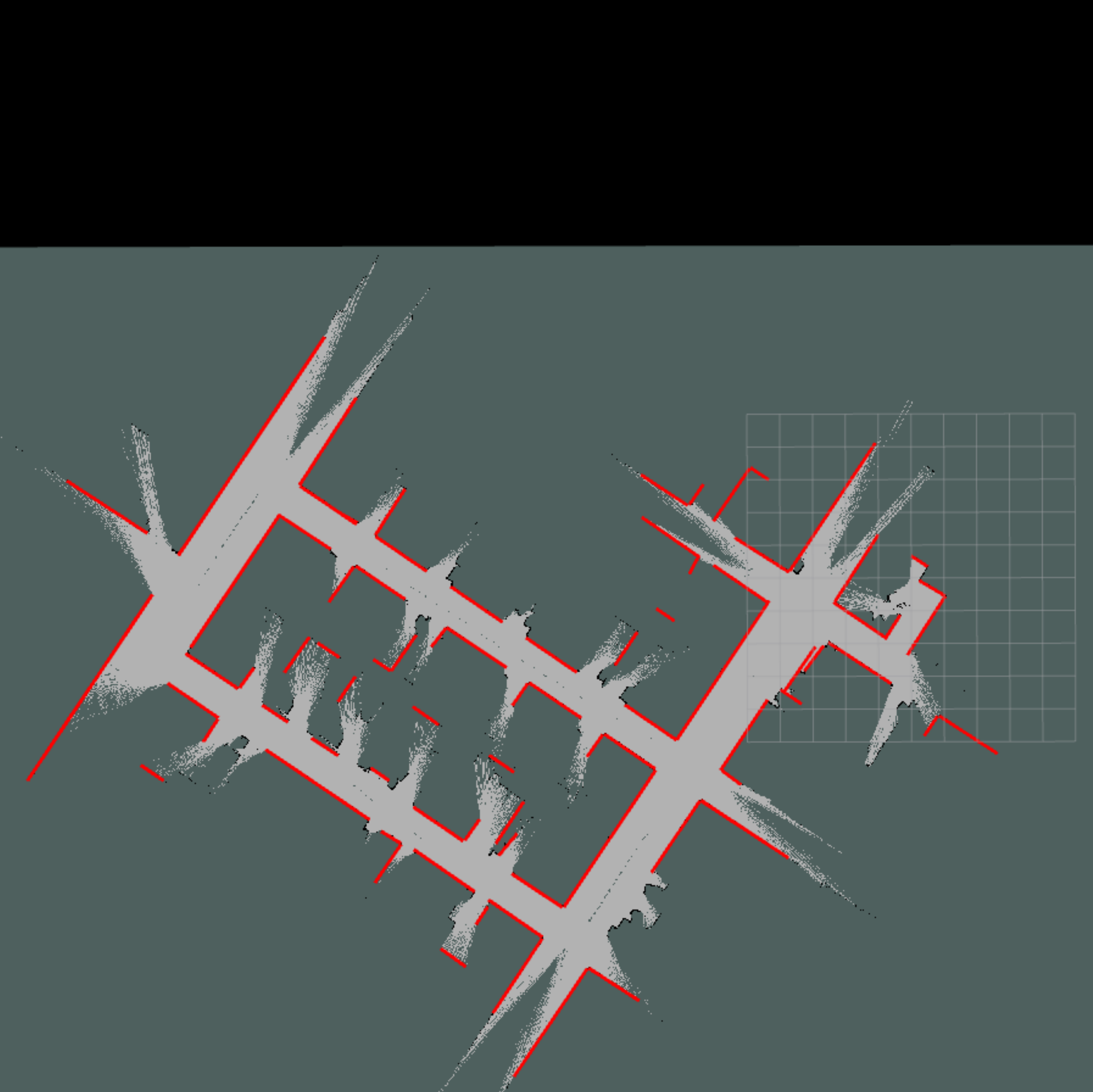}}
    \centering
    \subfigure[]{\includegraphics[scale=0.09]{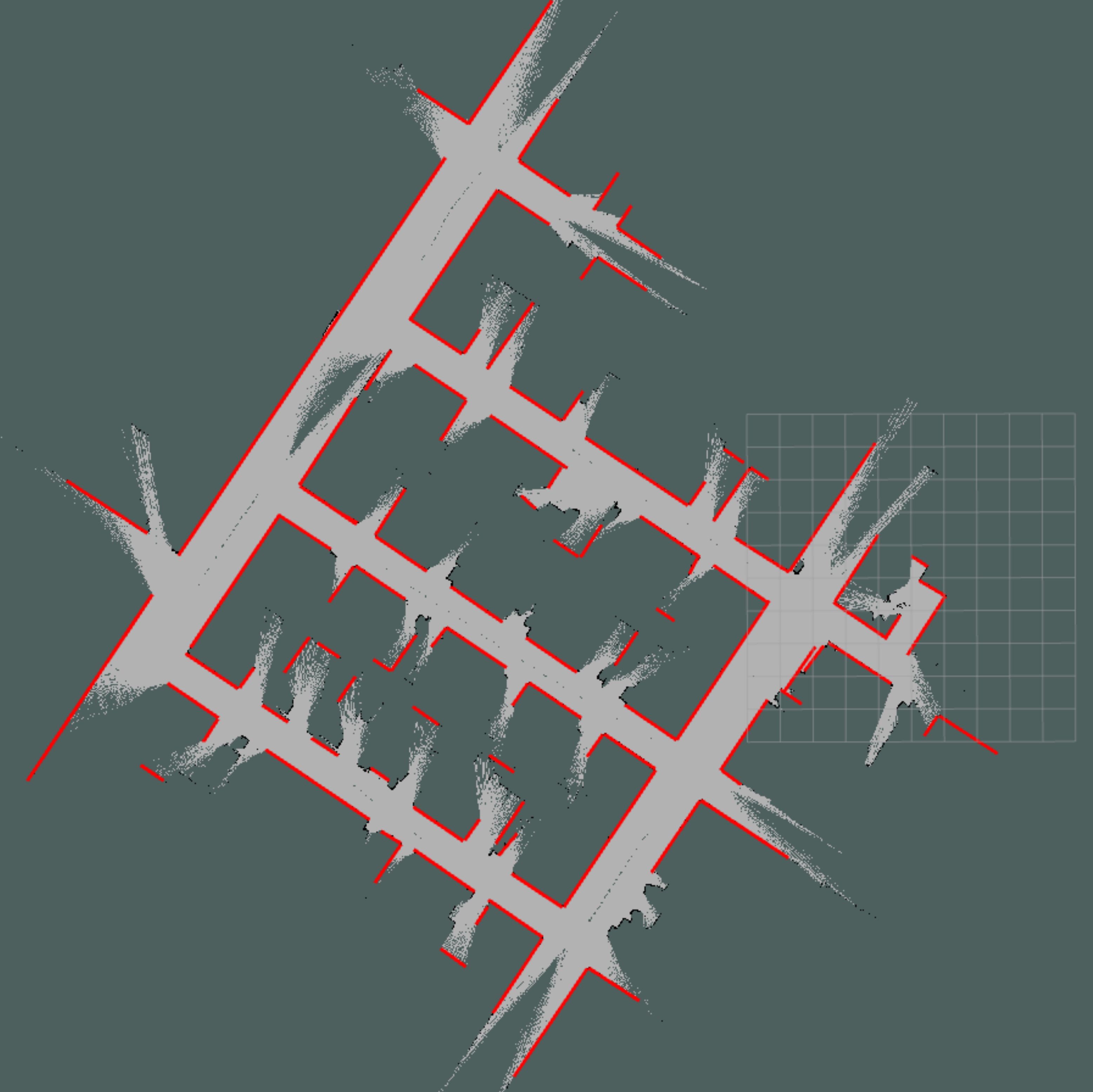}}
    \centering
    \subfigure[]{\includegraphics[scale=0.09]{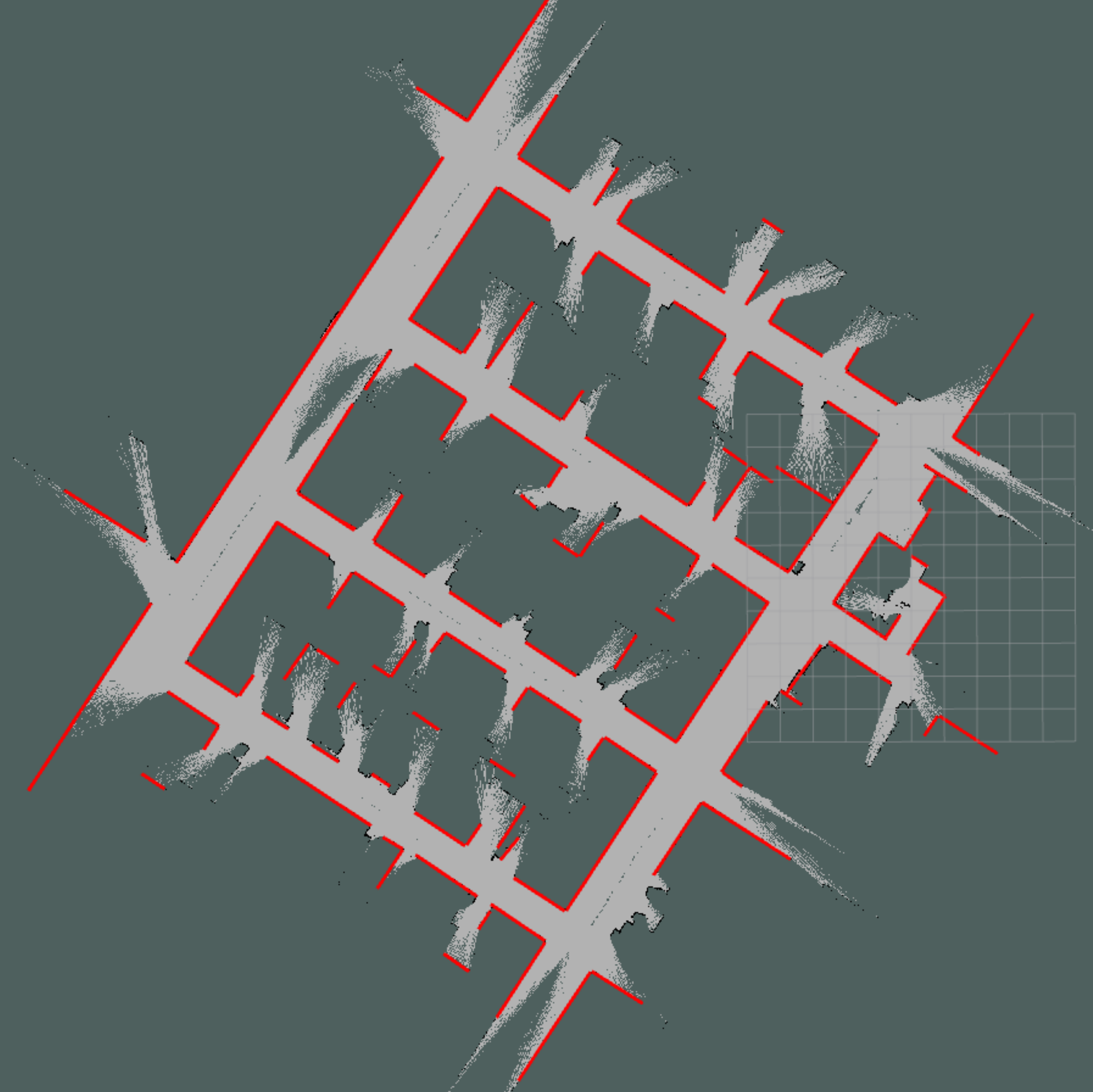}}
    \centering
    \subfigure[]{\includegraphics[scale=0.09]{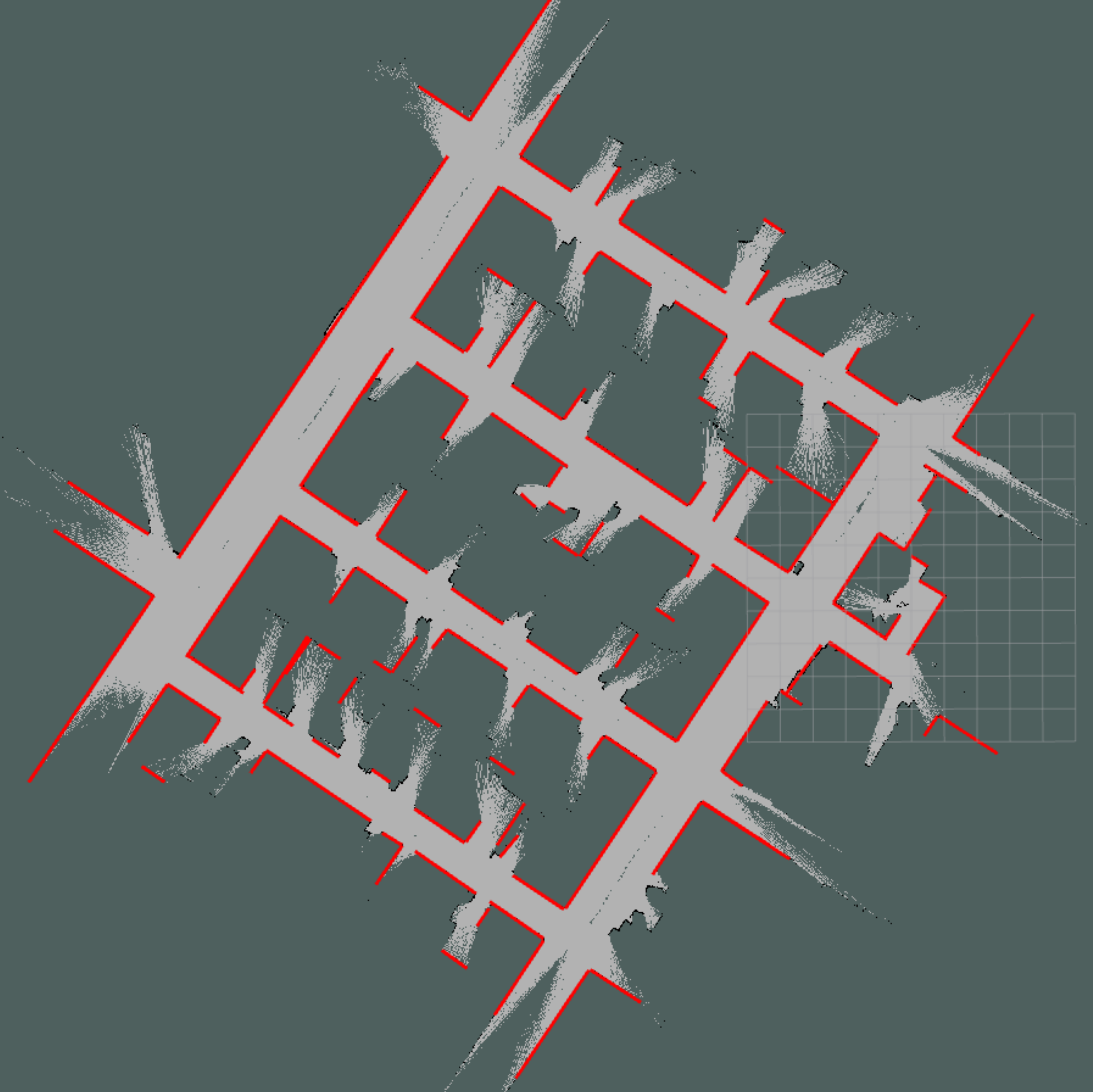}}
    \caption{Screenshot of the real-time mapping performance of CAE-RLSM on dataset (b). The red line segments represent the global map, and the grid map is obtained by Karto.}
    \label{real_time}
\end{figure}

\begin{figure}[t]
    \centering
    \subfigure[]{\includegraphics[scale=0.17]{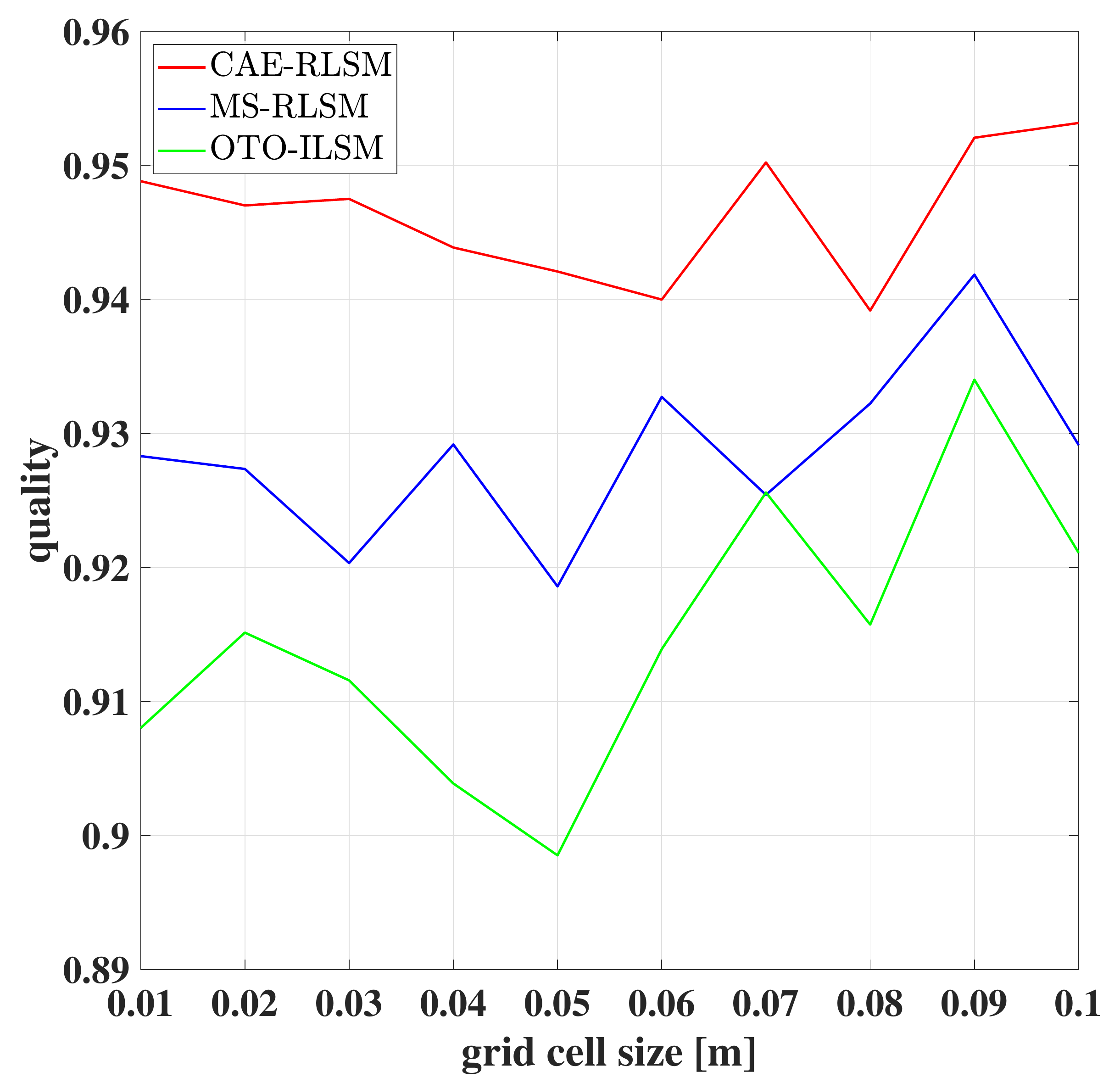}}
    \centering
    \subfigure[]{\includegraphics[scale=0.17]{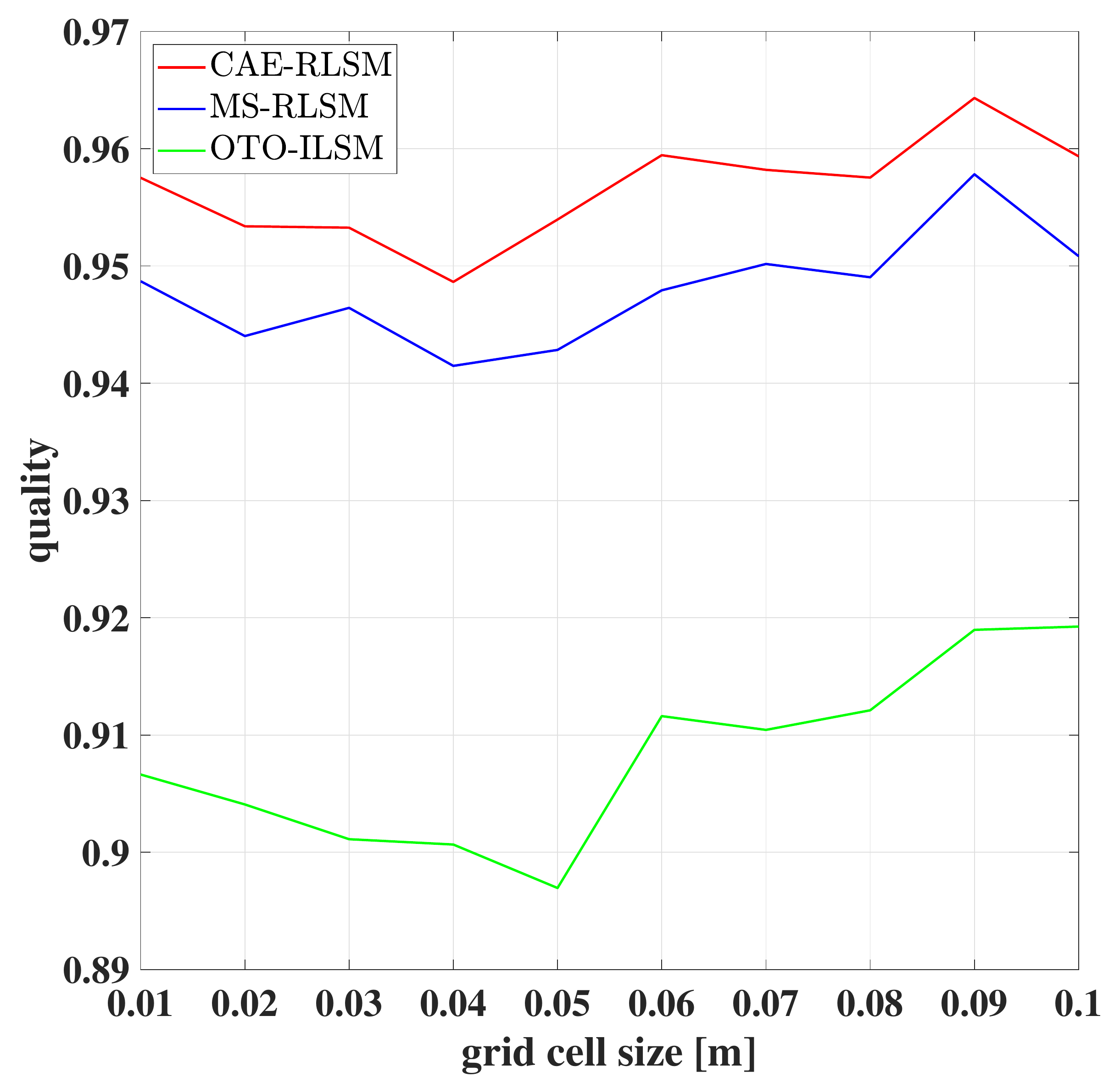}}
    \centering
    \subfigure[]{\includegraphics[scale=0.17]{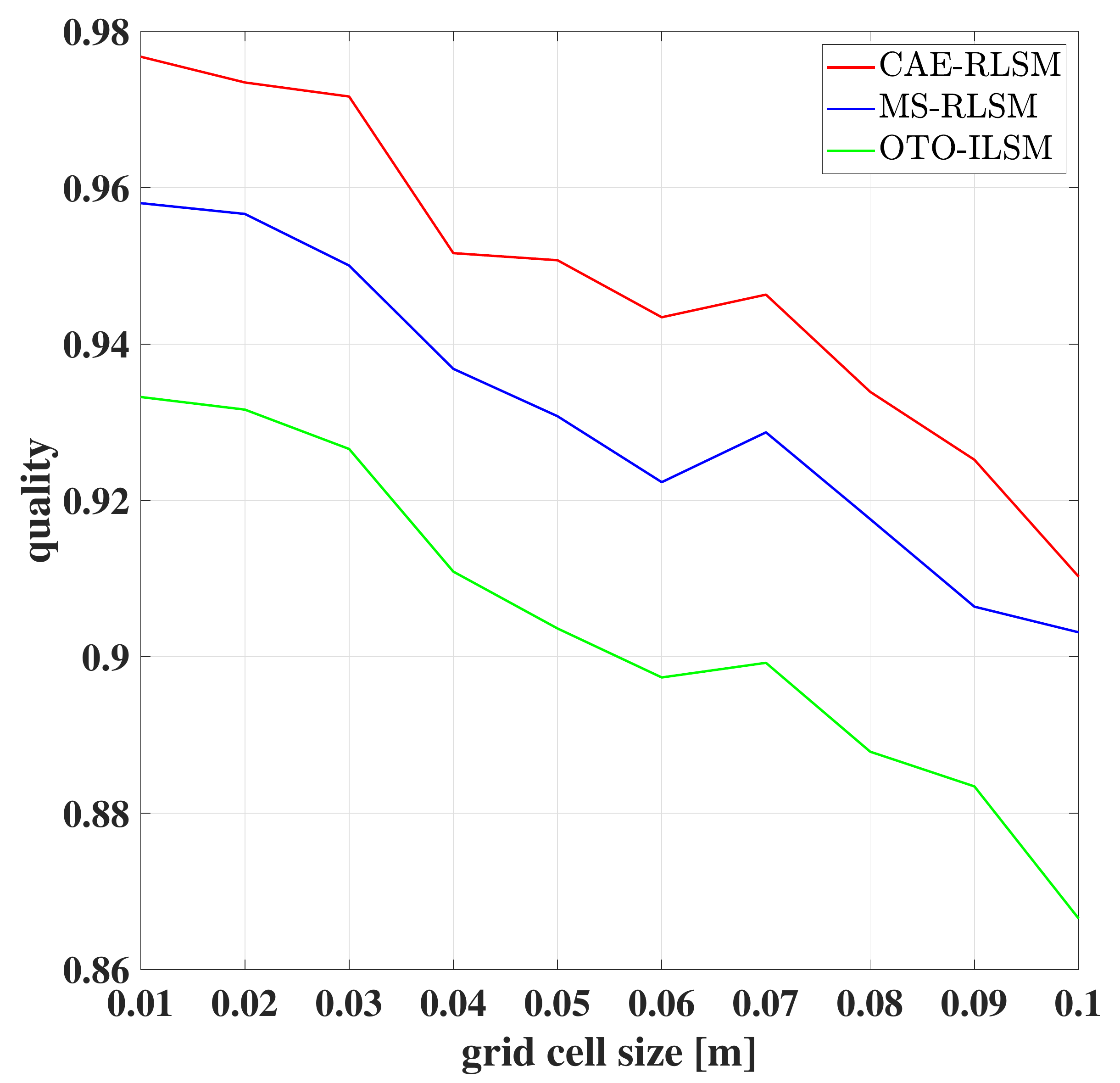}}
    \centering
    \subfigure[]{\includegraphics[scale=0.17]{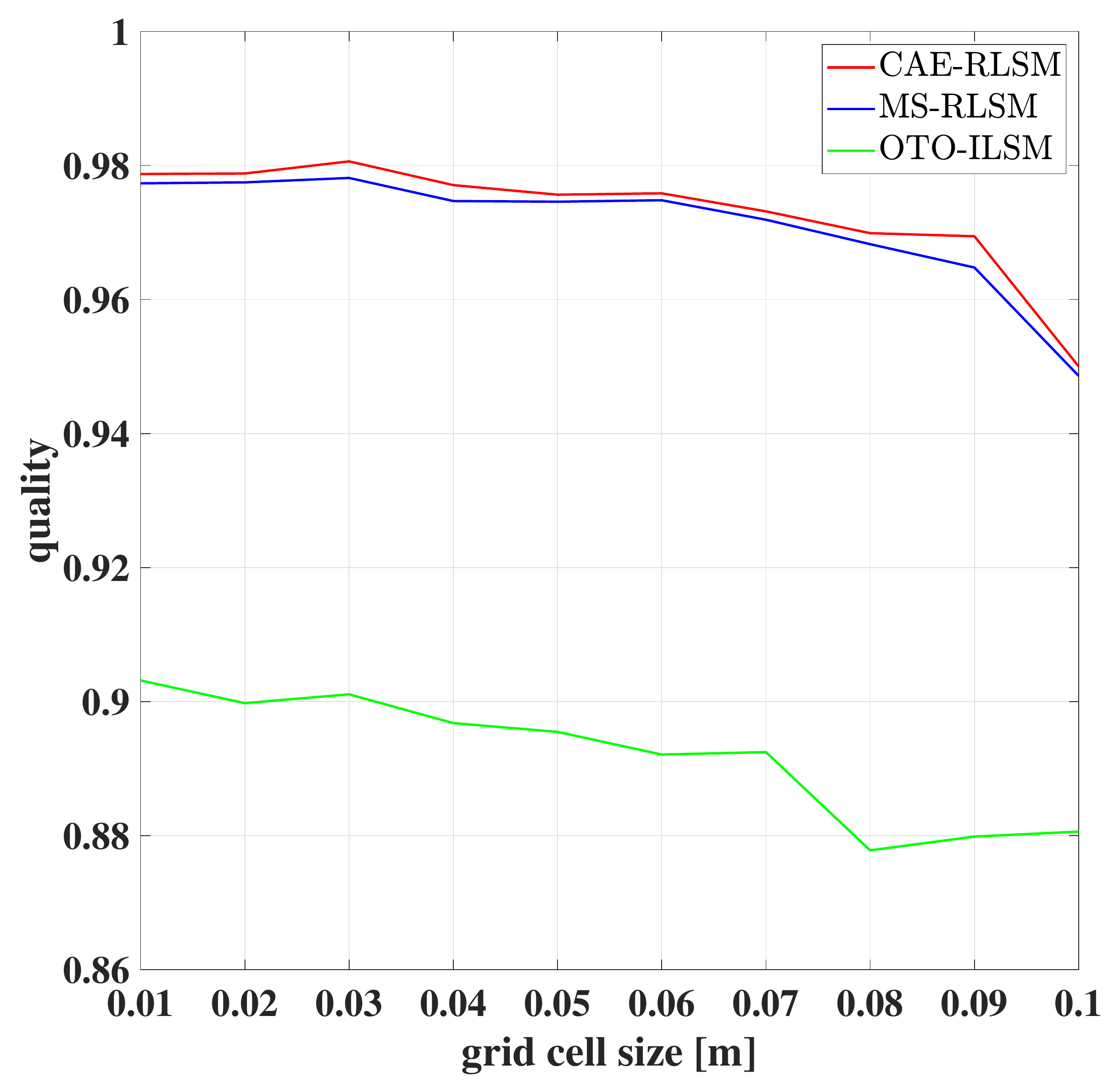}}
    \caption{Comparative experimental results of map quality with changing grid cell sizes on datasets (a) to (d).}
    \label{gridcellsize}
\end{figure}

Table \ref{table_3} enumerates a few attributes of the final line segment maps produced by O$^{2}$TO-ILSM. Compared with O$^{2}$TO-ILSM, the map quality of CAE-RLSM increases by an average of 5.8\%, the error reduces by 2.12 mm on average and the distance deviation reduces by an average of 0.94 mm. The final line segment maps produced by O$^{2}$TO-ILSM and CAE-RLSM on dataset (d) are used to intuitively compare the performance of these two approach. As shown in the Fig. \ref{floor_4}(c) and Fig. \ref{floor_4}(d), there are still redundant line segments in the final map produced by O$^{2}$TO-ILSM, while using CAE-RLSM can obtain a non-redundant line segment map. Furthermore, the low map quality of O$^{2}$TO-ILSM is consistent with the redundancy of the obtained map, which demonstrates that the proposed evaluation metric can directly reflect the redundancy of line segment maps.

In summary, the correlation-based metric compares the final line segments with the registered laser scan points while the error metric and distance metric compare the final line segments with the corresponding original line segments. It should be noted that both the laser scan points and original line segments contain the actual environmental information. It means that the correlation-based metric, error metric and distance metric all evaluate the coincidence degree between the final line segment map and the environment, the former is in discrete space and the last two are in continuous Euclidean space. Therefore, the evaluation results based on above evaluation metrics should be consistent.

\subsubsection{Efficiency comparison}
Finally, the real-time mapping performance of CAE-RLSM is visualized with the aid of Robot Operating System (ROS) visualization tools (RViz), as shown in Fig. \ref{real_time}. The per-frame runtime results are indicated in the Per-frame runtime (ms) column of Table \ref{table_3}. Since the post-processing mechanism makes MS-RLSM essentially impossible to achieve real-time mapping performance, we cannot get and list the per-frame runtime of MS-RLSM. Therefore, we use ``N/A'' to show that the corresponding item is not available. The per-frame runtime of CAE-RLSM is approximatively 0.08 milliseconds on average, which demonstrates that CAE-RLSM can ensure high efficiency.

In addition, we design an offline version for CAE-RLSM like O$^{2}$TO-ILSM to compare the total runtime results with respect to MS-RLSM and O$^{2}$TO-ILSM. The experimental results are shown in the Total offline runtime (ms) column of Table \ref{table_3}. The total runtime results of MS-RLSM all exceed 5 seconds (except for dataset (c), which contains only 240 laser scans), while the total runtime results of O$^{2}$TO-ILSM and CAE-RLSM are all less than 80 milliseconds. As can be seen, the difference of runtime results between MS-RLSM and CAE-RLSM is two orders of magnitude (except for dataset (c)), while the runtime results of O$^{2}$TO-ILSM and CAE-RLSM are the same order of magnitude.

\subsubsection{Practical considerations}
The correlation-based metric is a resolution-dependent metric. In order to investigate the relationship between the map quality and changing grid cell sizes, we test the grid cell size in the range of 0.01 m to 0.10 m with the step size of 0.01 m. The comparative experimental results of map quality with changing grid cell sizes on datasets (a) to (d) are presented in Fig. \ref{gridcellsize}. Since the registered laser scan points and discrete line segment pixels will hit in different grid cells according to different grid cell sizes, the change of the map quality with changing grid cell sizes is not regular. However, as shown in Fig. \ref{gridcellsize}, CAE-RLSM always achieves the best performance among these three approaches under the evaluation of the correlation-based metric with the same given grid cell size.

\section{Conclusion}
This work has investigated the problem of redundant line segment merging. A consistent and efficient redundant line segment merging approach called CAE-RLSM for online feature map building is proposed, which is composed of one-to-many incremental line segment merging (OTM-ILSM) and multi-processing global map adjustment (MP-GMA). The proposed CAE-RLSM not only reduces the redundancy of incremental merging, but also solves the problem of global map adjustment after loop closing, simultaneously ensuring efficiency and global consistency. Furthermore, a new correlation-based metric is proposed for quality evaluation of line segment maps. Comparative experimental results demonstrate that the proposed CAE-RLSM can achieve superior performance in terms of efficiency and map quality in different scenarios.

However, in this work we do not discuss in depth about the endpoint generation after redundant line segment merging. Actually, two line segments connecting a corner should be regarded as one endpoint. This issue may be called redundant endpoint elimination. Future work also includes application of the proposed approach to the real-time autonomous navigation systems.

\section{Acknowledgments}
The public datasets used in the experiments are obtained from the Robotics Data Set Repository (Radish) \cite{15}. The authors gratefully thank Gian Maria Pelusi, Maxim Batalin and Dirk Haehnel for providing the Department of DIIGA, the Intel Lab in Oregon and the Seattle datasets. The self-recorded dataset is collected on the fourth floor of the College of Artificial Intelligence, Nankai University. Thanks go to Lei Zhou and Meng Liu for providing this dataset.

\end{document}